\documentclass[twocolumn]{svjour3}          
\smartqed 
\usepackage{graphicx}
\usepackage[numbers]{natbib}
\usepackage{times}
\usepackage{epsfig}
\usepackage{amsmath}
\usepackage{amssymb}
\usepackage{lineno}
\usepackage{pifont}
\usepackage{longtable}
\usepackage{supertabular}

\usepackage{adjustbox}

\usepackage[pagebackref=true,breaklinks=true,letterpaper=true,colorlinks,bookmarks=false,citecolor=blue,linkcolor=blue]{hyperref}

\usepackage{booktabs}
\usepackage{amsmath}
\usepackage{amssymb}
\usepackage{verbatim}
\usepackage{multirow, makecell}
\usepackage{lineno}
\usepackage{enumerate}
\usepackage{amsfonts}
\usepackage{gensymb}
\usepackage{bm}
\usepackage{bbding}
\usepackage{comment}
\usepackage{color}
\usepackage{diagbox}
\usepackage{tabularx}
\usepackage{rotating}
\newcolumntype{L}[1]{>{\raggedright\let\newline\\\arraybackslash\hspace{0pt}}m{#1}}
\newcolumntype{C}[1]{>{\centering\arraybackslash}m{#1}}
\newcolumntype{R}[1]{>{\raggedleft\let\newline\\\arraybackslash\hspace{0pt}}m{#1}}

\newlength\savewidth

\usepackage{colortbl}
\definecolor{Gray}{gray}{0.9}
\definecolor{GrayT}{gray}{0.4}

\usepackage{subcaption}
\usepackage[labelfont=bf]{caption}
\definecolor{light-blue}{RGB}{186,175,255}
\definecolor{light-red}{RGB}{255,137,165}
\usepackage{arydshln} 
\usepackage{booktabs}
\usepackage{multirow}
\usepackage{amsmath}
\usepackage{etoolbox}
\usepackage{bm}
\usepackage{mathtools}
\usepackage{algorithm}
\usepackage{algpseudocode}
\usepackage{xspace}
\usepackage{longtable}

\usepackage{multirow, makecell}
\usepackage{colortbl}  
\usepackage{xcolor}
\usepackage{array}
\definecolor{Gray}{gray}{0.94}
\definecolor{liGray}{gray}{0.5}
\definecolor{LightCyan}{rgb}{0.88,1,1}
\usepackage{mathrsfs}

\definecolor{dgreen}{rgb}{0.0,0.6,0.0}
\newcommand{\dataset}{LvBench\xspace}
\newcommand{\method}{DPM\xspace}
\newcommand{\cmark}{\textcolor{dgreen}{\ding{51}}}
\newcommand{\xmark}{\textcolor{red}{\ding{55}}}
\definecolor{darkblue}{RGB}{46,25, 110}

\begin{document}
\begin{sloppypar}

\title{\dataset: A Benchmark for Long-form Video Understanding with Versatile Multi-modal Question Answering}

\author{Hongjie Zhang$^{1*}$      \and
        Lu Dong$^{1,2*}$    \and
        Yi Liu$^{3*}$      \and
        Yifei Huang$^{1}$       \and
        Yali Wang$^{1,4}$       \and
        Limin Wang$^{1,5}$\textsuperscript{†}  \and 
        Yu Qiao$^{1}\textsuperscript{†}$ 
}

\institute{Hongjie Zhang \at
              \email{nju.zhanghongjie@gmail.com}
           \and
           Lu Dong \at
              \email{dl1111@mail.ustc.edu.cn}
           \and
           Yi Liu \at
              \email{yi.liu1@siat.ac.cn}
           \and
           Yifei Huang \at
              \email{hyf015@gmail.com}
           \and
           Yali Wang \at
           \email{yl.wang@siat.ac.cn}
           \and
           Limin Wang \at
           \email{lmwang@nju.edu.cn}
           \and
           Yu Qiao \at
           \email{yu.qiao@siat.ac.cn}
           \\
           $^1$ OpenGVLab, Shanghai AI Laboratory, China
           \\
           $^2$ University of Science and Technology of China, China
           \\
           $^3$ Honor Device Co.,Ltd
           \\
           $^4$ Shenzhen Institutes of Advanced Technology, Chinese Academy of Sciences, China
           \\
           $^5$ Nanjing University, China
           \\
}

\maketitle

\let\oldthefootnote\thefootnote
\let\thefootnote\relax
\footnotetext[0]{\hspace*{-1.8em}%
    \small $^*$ Equal contribution;\quad $^\dagger$ Corresponding Authors}
\let\thefootnote\oldthefootnote

\begin{abstract}
Despite remarkable recent progress, existing long-form VideoQA datasets fall short of meeting the criteria for genuine long-form video understanding. This is primarily due to the use of short videos for question curation, and the reliance on limited-length sub-clips as clues to answer those questions. Meanwhile, previous datasets have limited focus on question type and modality.
To remedy this, we introduce \textbf{\dataset}, 
a \textbf{L}ong-form \textbf{V}ideo understanding \textbf{Bench}mark for versatile multi-modal question-answering.
Our \dataset stands out from existing long-form VideoQA datasets through three key characteristics:
\textbf{\textit{1) Extended temporal durations}}: We consider videos ranging from 70 seconds to 4 hours, covering single-scene, multi-scene, and full-scene contexts. This design accounts for both video and clue lengths, capturing diverse contextual dynamics.
\textbf{\textit{2) Diverse question types and modalities}}: \dataset introduces six distinct question types that evaluate various perceptual and cognitive capabilities, utilizing both video frames and subtitles.
\textbf{\textit{3) High-quality annotations}}: We employ rigorous manual labeling by human annotators. Our dataset comprises 20,061 question-answer pairs sourced from 100 carefully selected movies across diverse genres, annotated collaboratively by multiple individuals.
Analysis involving various baselines reveals a consistent trend: the performance of all existing methods significantly deteriorates when video and clue length increases. 
We expect \dataset to serve as a valuable resource for future works on long-form video understanding.
\end{abstract}

\vspace{-1em}
\section{Introduction}
\label{sec:intro}

\begin{figure*}[t]
\begin{center}
\includegraphics[width=0.95\linewidth]{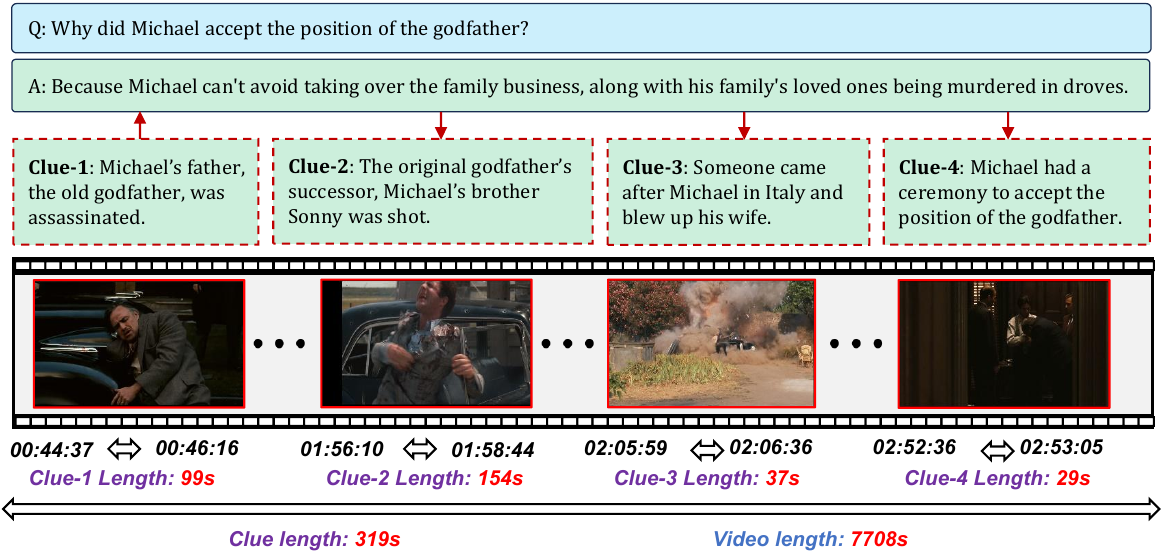}
\end{center}
\caption{A causal reasoning QA example from \dataset about the movie \emph{The Godfather}. Answering this question not only requires capturing clues from a long temporal span but also requires high-level processing of these relevant clues.}
\label{fig:exemplar}
\end{figure*}

Long-form videos, with their intricate narratives and extended duration, offer a rich source of information and a unique challenge for understanding. 
To truly understand their content, one must piece together the various clues scattered throughout the video ~\cite{maharaj2017dataset}. 
Movies serve as an illustrative example of the challenges inherent in long-form video understanding~\cite{wu2021towards}. As shown in Fig.~\ref{fig:exemplar}, answering a question in the movie \emph{The Godfather} requires piecing together various clues from the long storyline. These clues may be located tens of minutes away from the specific clip where the question is posed.

While humans can easily capture these clues and reason about their causal and temporal relations~\cite{argaw2023long}, it remains a great challenge for current multi-modal large language models (MLLM) to demonstrate comparable capabilities. 
We believe this discrepancy stems from the absence of a suitable benchmark. Existing video question answering (VideoQA) benchmarks fall short in genuinely addressing long video understanding~\cite{castro2020lifeqa,yu2019activitynet,xiao2021next,wu2021star}, primarily due to their limitations in two crucial aspects: clue length and video length. The clue length, defined as the minimum sub-clip set required to verify annotated information, is a crucial indicator of the inherent temporal difficulty of long-form VideoQA tasks. On the other hand, longer videos introduce more information redundancy and complex temporal relationships, further increasing the overall difficulty. Therefore, 
both clue length and video length play essential roles in accurately assessing the intricacies of long video understanding.

\begin{figure*}[t]
\begin{center}
\includegraphics[width=\linewidth]{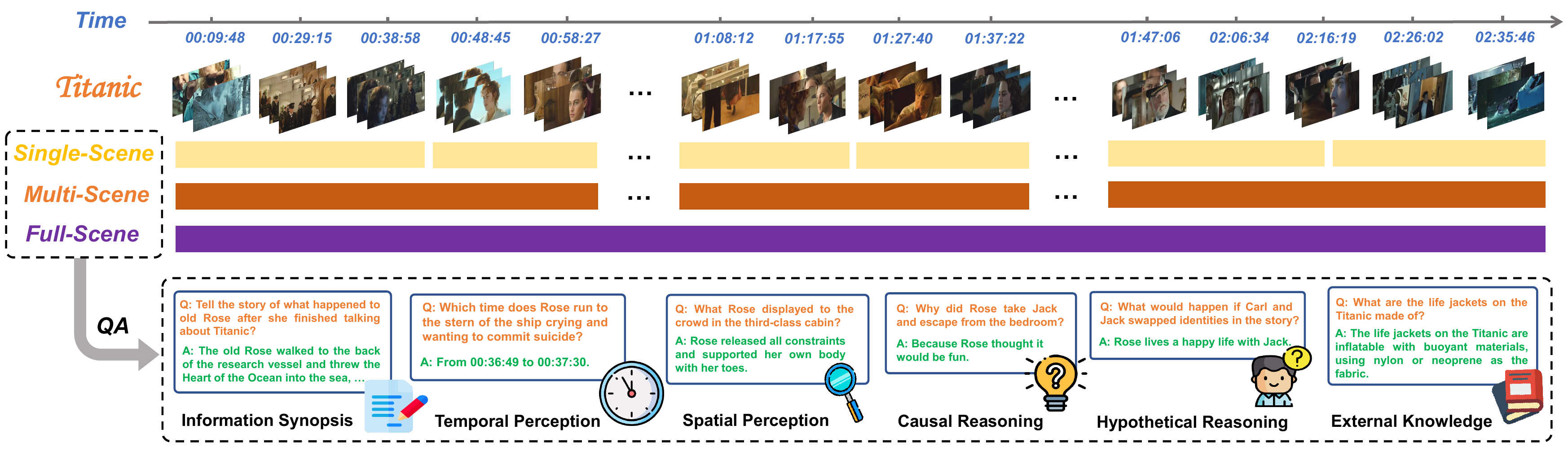}
\end{center}
\caption{An overview of our \dataset.  We use multi-level question-answering to systematically evaluate the long-form video understanding capabilities of existing models. These QA pairs are sourced from single-scene, multi-scene, and full-scene, and cover various aspects of video understanding, including information synopsis, temporal perception, spatial perception, causal reasoning, hypothetical reasoning, and external knowledge.
}
\label{fig:framework}
\end{figure*}

In this work, considering both indicators, we introduce \dataset, a new benchmark containing 20,061 manually annotated QA pairs sourced from 100 movies with diverse genres, to assess the model capabilities on long-form video understanding. To enable a comprehensive evaluation, 
we separate our QAs based on their video lengths into three levels: single-scene, multi-scene, and full-scene (Fig.~\ref{fig:framework}). The scenes are manually partitioned based on the video content, with lengths ranging from 70 seconds to 4 hours. This video length and the resulting clue length (post-evaluated) significantly surpass existing VideoQA benchmarks such as~\cite{mangalam2023egoschema,song2023moviechat,tapaswi2016movieqa}. In addition to the long video and clue lengths, QAs in \dataset are designed from the perspective of the interconnected abilities required by moviegoers to understand the movie content. We specifically design six types of QA: information synopsis, temporal perception, spatial perception, causal reasoning, hypothetical reasoning, and external knowledge. Unlike previous works that solely rely on textual plots~\cite{tapaswi2016movieqa} or vision clues~\cite{mangalam2023egoschema}, our dataset encompasses multi-modalities, to capture the multifaceted nature of movie understanding. With our careful ensure of video/clue lengths and rigorous annotation, we believe \dataset can serve as a valuable resource to incentivize the development of multimodal systems that can tackle versatile QAs spanning long periods and relying on multiple modality information.

To summarize, the main contributions of this work are:
(1) We introduce \dataset, a benchmark specifically tailored for long-form video understanding, taking into account both long video length and clue length as key factors.
(2) We design and annotate multiple types of QA that encompass various aspects of video understanding. Additionally, our QAs consider multiple modalities, enabling the assessment of model capabilities on various perceptual and cognitive axes.
(3) We benchmark various state-of-the-art multimodal systems on \dataset and find that even the most advanced current systems achieve unsatisfactory results. Meanwhile, we establish a simple dual-pathway framework which we hope can serve as an initial step in establishing long-form video understanding methods.

\section{Related Work}
\label{sec:related}
\subsection{Video Question-Answering Datasets}
VideoQA is a powerful tool for evaluating a model's comprehensive understanding of a given video \cite{mun2017marioqa, jang2017tgif, maharaj2017dataset, xu2017video, castro2022wildqa,choi2021dramaqa,tapaswi2016movieqa,lei2018tvqa,lei2019tvqa+,xiao2021next}. Early VideoQA datasets, such as TGIF-QA \cite{jang2017tgif}, 
MSRVTT-QA \cite{xu2017video} primarily focus on visual descriptions.
Recent advancements in VideoQA aim for more causal and temporal reasoning on multiple events. NExT-QA \cite{xiao2021next} emphasizes causal and temporal actions in daily life videos. Datasets like MovieQA \cite{tapaswi2016movieqa} and TVQA \cite{lei2018tvqa} design causal questions on movies and TV shows respectively, requiring plot understanding or actor dialogue understanding. Nevertheless, these datasets ask questions on relatively short videos (ranging from 10 to 203 seconds), while our \dataset asks questions on much longer videos (948 seconds on average) with much richer video content. 

\subsection{Long-form Video Understanding Datasets}
Recently, long-form video understanding has begun to rise \cite{mullapudi2019online,huang2020improving,wu2021towards,soldan2022mad,mangalam2023egoschema,shou2021generic,huang2020movienet,vicol2018moviegraphs,xiong2019graph,yue2023movie101,grauman2022ego4d,huang2024egoexolearn}. \cite{wu2021towards} proposes a long-form video understanding benchmark based on \cite{movieclips}, where the tasks are much specific, such as speaker style and  release year.
\cite{soldan2022mad} proposes MAD, a language grounding dataset designed to ground queries on an average clip duration of 110 minutes. However, their language queries are much shorter (averaging 4 seconds).
MovieChat-1k \cite{song2023moviechat} generates QAs for movie understanding, with mainly single-frame breakpoint questions,  lacking long temporal context understanding.
\cite{mangalam2023egoschema} proposes EgoSchema with a clue length of about 100 seconds.
However, their LLM-generated QAs lack diversity, mainly focusing on overall and primary events. In contrast, our manually labeled QAs with long clue lengths are more diverse and require an understanding of long temporal context. The average video length in \cite{rawal2024cinepile} is only 160 seconds, which is significantly shorter than the 948 seconds in our \dataset.
\cite{fu2024video,wang2024lvbench,wu2024longvideobench} introduce evaluation benchmarks of  Multi-modal Large Language Models in Video analysis, however, the number of QAs in these datasets is sufficient only for evaluation, not for training. In contrast, our \dataset includes 20,061 manually annotated multiple-choice questions, sufficient for model training.

\subsection{Long-form Video Understanding}
The main challenge in long-form video understanding is how to handle long video inputs~\cite{wu2021towards,han2022temporal,sun2022long,xiao2022hierarchical,wu2022memvit,zhang2016video}. There are two main directions. The first approach is to long-form modeling
\cite{huang2018predicting, wu2019long, lin2022eclipse, shou2016temporal}. Specifically, \cite{wu2019long} uses the non-local network to attend the long video features with short video feature query. \cite{lin2022eclipse} models long video using sparsely sampled high-cost video frames and dense sampled low-cost audio.  Another approach is to select keyframes of long videos~\cite{yang2021stacked,pei2017temporal,korbar2019scsampler,huang2023weakly,liu2022empirical,gao2023mist, potapov2014category}. \cite{gao2023mist} utilizes cascaded segment and region selection modules to select question-related  frames and image
regions. \cite{yu2023self} uses grounding datasets for keyframe localizer pretraining and leverages QA feedback to refine the keyframe localizer. We combine the merit of existing works to design a dual-pathway modeling framework for efficient long video understanding. 

\section{\dataset}
\label{sec:benchmark}
We carefully select 100 movies from the top-rated movies on IMDb~\cite{imdb}, covering a wide range of genres, years, and countries ((see Appendix~\ref{appx:100movies} for their statistics)). The average length of the movies is 125.7 minutes.
To ensure the purity of the visual information in our experiments, the movie data we used does not include embedded subtitles. However, we have collected external subtitles for each movie, which provide the precise start and end timestamps for the dialogue in the video content.
We engaged the expertise of 50 annotators to ensure high-quality QA annotations. 
During the annotation process, we temporarily included embedded subtitles in the movie data to assist the annotators in understanding the movie content. Below, we describe in detail our QA design and annotation process.

\subsection{Question-Answer Design} \label{qa_collection}

\noindent\textbf{Scene Partition.}
To evaluate the capabilities of models in understanding long-form videos with varying temporal lengths, we employ a segmentation process for each movie in our dataset. Annotators first select movies that they are familiar with from a provided list. They then manually segment these movies into consecutive, non-overlapping single-scene intervals. The segmentation is based on the storyline and visual presentation, resulting in clear start and end timestamps for each scene. On average, the duration of a single scene is approximately 7.8 minutes.
For longer video understanding, we merge adjacent and closely related single-scenes to create multi-scenes with extended durations. These multi-scenes typically span an average length of 22 minutes. Additionally, our dataset includes specifically designed questions and answers for super long-form understanding, focusing on the full-scene duration of the movies, which averages 125.7 minutes.

\noindent\textbf{QA Annotation.} 
We design QAs for each segmented scene to assess the comprehension of models across different temporal spans.
Annotators are given the freedom to formulate 6 to 12 questions per scene, covering 6 different types. These questions require both vision and language understanding to be answered effectively.
To mitigate subjective biases that may arise from individual annotators, we employ a collaborative annotation approach. Each movie is annotated by at least two annotators, with one responsible for asking the questions and the other providing the answers while also correcting any mistakes. This ensures a more objective and reliable annotation process. Furthermore, to enhance the annotation quality, two additional reviewers thoroughly checked the annotated QAs. They were incentivized to identify any mistakes, which would result in a cash bonus. 

\noindent\textbf{QA Types.} 
To assess model capabilities across different perceptual and cognitive dimensions in long-form video understanding, our dataset includes a range of question types that are designed from the perspective of the interconnected abilities required by humans to comprehend video content. They are grouped into six types:

\noindent$\bullet$~\textit{Information Synopsis.} Information synopsis questions require the model to extract key information related to the question from the movie and summarize them concisely in language. This involves aspects such as the movie theme, plot synopsis, character relationships, and motivations. 

\noindent$\bullet$~\textit{Temporal Perception.} Temporal perception questions require the model to
understand the timeline of events and the plot in the movie, and accurately locate the temporal location of the answer on the timeline.
Compared with the temporal action localization task~\cite{zhang2022actionformer,cheng2022tallformer}, which requires the model to recognize and locate specific actions in a video, our temporal perception task places more emphasis on understanding the complex narrative sequence in the movie. 

\noindent$\bullet$~\textit{Spatial Perception.} Spatial perception questions require the model to recognize and understand the visual elements in movies, which often have multiple characters, objects, and scenes. It may require combining the context of the story to infer the visual information related to the question.

\noindent$\bullet$~\textit{Causal Reasoning.} Causal reasoning questions require the model to understand and infer cause-and-effect relationships between events in the storyline, and causality is the key driving force in the development of the story. 

\noindent$\bullet$~\textit{Hypothetical Reasoning.} Hypothetical reasoning questions require the model to understand the preconditions and post-effects of storylines. Based on the causal tracing of storylines, we designed hypothetical reasoning questions with the assumption that certain events in the story do not happen, and how the storyline will develop under this condition.

\noindent$\bullet$~\textit{External Knowledge.} External knowledge questions require the model to understand additional knowledge that could not be found in movies but has relevance to the movie content. This external knowledge may involve cultural background, actor information, history, and social response. 

Examples of the various types mentioned above are showcased in Appendix~\ref{appx:datasets_examples}. We also present the ground truth answers and the results predicted by our model.

\subsection{Data Statistics}
\begin{figure*}[t]
\begin{center}
\includegraphics[width=\linewidth]{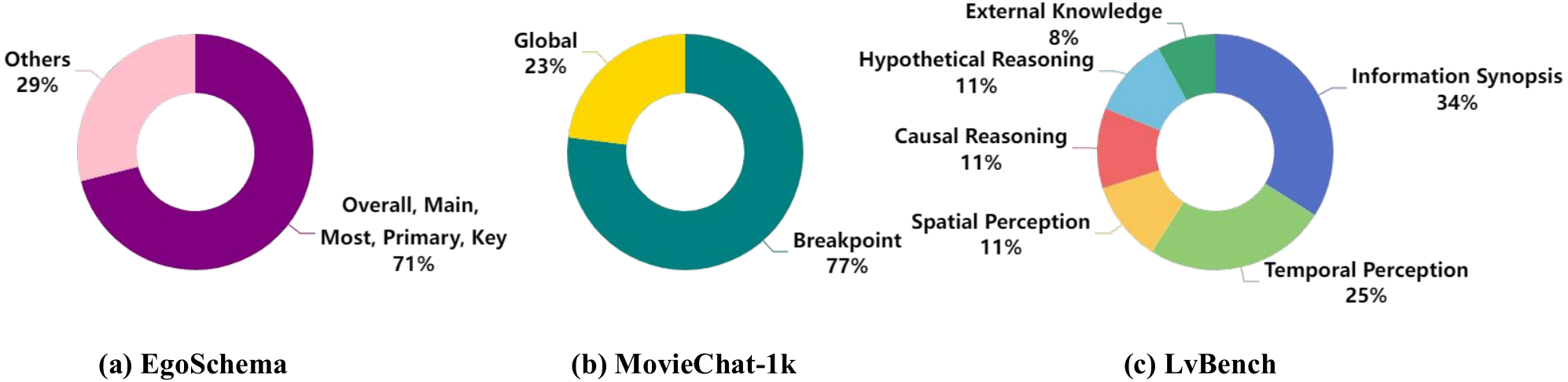}
\end{center}
\caption{ 
The distribution of (a) Keywords in EgoSchema, (b) Question modes in MovieChat-1k, (c) QA types in \dataset. This shows that EgoSchema and MovieChat-1k are both biased towards a certain type of questions. Best viewed in color.
}
\label{fig:q_type_comparison}
\end{figure*}

We collect a total of 20,061 QAs, including 14,617 single-scene, 4,635 multi-scene, and 809 full-scene. 
The distribution of different types of QAs in our \dataset is shown in Fig.\ref{fig:q_type_comparison} (c).
The number of QAs in three scenes accounted for 73\% in single-scene, 23\% in multiple-scene, and 4\% in full-scene.
We follow ~\cite{mangalam2023egoschema} and benchmark the clue length for 50 hours of clips (200 QAs) chosen randomly. 
Our \dataset has a median clue length of about 200 seconds for single-scene, 320 seconds for multi-scene, and 540 seconds for full-scene, which are $2 \times$ (single-scene), $3.2 \times$ (multi-scene), and $5.4 \times$ (full-scene) longer than the dataset with the second-longest clue length.
In general, the median clue length of all QAs is approximately 230 seconds.
Additionally, our \dataset has an average video length of about 15.8 minutes for all QAs, which is $4.7 \times$ longer than the second-longest video length. 

Table.~\ref{tab:words} presents statistics of the QAs based on types. For the number of words, on average, both questions and answers in our \dataset have a higher word count compared to previous datasets, especially the answers, which are approximately $3 \times$ longer than those in TVQA~\cite{lei2018tvqa} and MovieQA~\cite{tapaswi2016movieqa}.
This indicates that our dataset provides richer and more detailed information, requiring more clues to describe and answer the questions.

\begin{table}
\caption{Statistics for different QA types. Q = question, CO = correct option, DO = distractor option. Length is defined as the number of words.}
\label{tab:words}
\centering
\scalebox{1}{
\begin{tabular}{c|ccc}
\toprule
Q Type  & Q-Len & CO-Len & DO-Len \\
\midrule
Information Synopsis  & 11.0 &  28.9 & 28.7 \\ 
Temporal Perception   & 12.1  & 3.0 & 3.0 \\
Spatial Perception    &  10.8 & 13.9 & 13.4 \\
Causal Reasoning      & 10.9 & 15.0 & 15.3 \\
Hypothetical Reasoning& 14.9  &  14.7 & 14.8 \\
External Knowledge    &  8.7 & 26.9 & 26.6 \\
\midrule
Total                 &  11.5 &  17.5 & 17.4 \\
\bottomrule
\end{tabular}
}
\end{table}

\begin{figure}[t]
\begin{center}
\includegraphics[width=\linewidth]{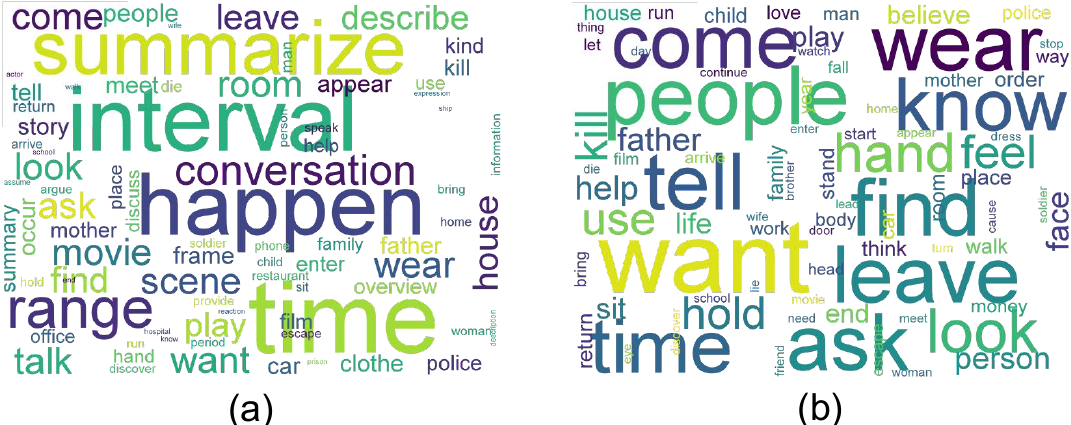}
\end{center}
\caption{Word cloud of the nouns/verbs in the questions and answers of  \dataset. (a): Question, (b): Answer.
}
\label{fig:wordcloud}
\end{figure}

Fig.~\ref{fig:wordcloud} shows the word cloud of the nouns/verbs in the questions and answers of our \dataset.
We can see that both the questions and answers cover various aspects and exhibit a high level of diversity in the word cloud. This diversity is also attributed to the fact that our question-answering system itself handles different types of questions.

\subsection{Comparison with Other Datasets}
In Table \ref{table:dataset}, we compare \dataset to several popular VideoQA datasets.  \dataset stands out in terms of video length, clue length, and the diversity of QA types. 
Specifically, we compare \dataset with five closely related benchmarks: MovieQA~\cite{tapaswi2016movieqa}, TVQA~\cite{lei2018tvqa}, EgoSchema~\cite{mangalam2023egoschema}, MovieChat-1k~\cite{song2023moviechat}, VideoMME~\cite{fu2024video} and Perception Test~\cite{patraucean2023perception}.
For MovieQA, there are two major differences. (1) \textit{Annotation method}: The questions in MovieQA are constructed from paragraph-level plot synopses, focusing primarily on text-based narrative content while lacking visual grounding~\cite{lei2018tvqa}. In contrast, our \dataset requires annotators to directly watch the video and consider both visual and textual elements, resulting in multimodally grounded and context-rich questions. (2) \textit{Temporal length}: The video and clue durations in MovieQA are relatively short (video: 203 seconds, clue: 30 seconds), even shorter than our single-scene level (video: 468 seconds, clue: 200 seconds), highlighting the greater temporal complexity of our \dataset.
For TVQA, the temporal scope is even more limited, with average video and clue durations of only 76 and 10 seconds, respectively. This further highlights the need for benchmarks like ours that better reflect the challenges of long-form temporal reasoning.
\begin{table*}[t]
\caption{Comparison of our \dataset to various VideoQA datasets. Explanation for QA Type. (C: Causal, H: Hypothetical, Sy: Synopsis, K: Knowledge, Sp: Spatial, T: Temporal)}
\label{table:dataset}
\centering 
\scalebox{0.9}{
\begin{tabular}{ccccccccccccc} 
\toprule
\multirow{2}[2]{*}{Dataset}  & \multirow{2}[2]{*}{Annotation}  & \multirow{2}[2]{*}{QAs}    &  \multirow{2}[2]{*}{Avg.Len.(s)} &  \multirow{2}[2]{*}{Clue.Len.(s)} & \multirow{2}[2]{*}{Multi-level} & \multirow{2}[2]{*}{Multimodal}   & \multicolumn{6}{c}{QA Type} \\
\cmidrule(lr){8-13}
& & & & & & & \small{C} & \small{H} & \small{Sy} &  \small{K} & \small{Sp}  & \small{T}  \\
\midrule
TGIF-QA~\cite{jang2017tgif}         &Auto & 165,165                 & 3          & 1  & \xmark      & \xmark   & \xmark & \xmark & \xmark & \xmark & \cmark &  \xmark  \\ 
MSRVTT-QA~\cite{xu2017video} & Auto & 243,690 & 15 & 1 & \xmark      & \xmark   & \xmark & \xmark & \xmark & \xmark & \cmark &  \xmark\\
How2QA~\cite{li2020hero} & Human & 44,007 & 60 & 2 & \xmark       & \xmark   
& \cmark & \xmark & \xmark & \cmark & \cmark &  
\cmark \\
NExT-QA~\cite{li2020hero} & Human & 52,044 & 44 & 5 & \xmark       & \xmark   
& \cmark & \xmark & \xmark & \xmark & \cmark &  \xmark \\
EgoSchema~\cite{mangalam2023egoschema}  & Auto     & 5,000     & 180  & 100                            & \xmark   & \xmark   &  \cmark & \xmark &  \cmark  & \cmark & \xmark & \xmark     \\
MovieQA~\cite{tapaswi2016movieqa} & Human  & 6,462                    & 203      & 30    & \xmark      &  \xmark   
& \cmark & \xmark  & \xmark  &   \cmark &  \cmark & \xmark
\\
TVQA~\cite{lei2018tvqa}   & Human          & 152,545               & 76      & 10     & \xmark      & \cmark  
& \cmark & \xmark  & \xmark  &   \cmark &  \cmark & \cmark
\\ 
MovieChat-1k~\cite{tapaswi2016movieqa} & Human  & 19,017                    & 564      & 90    & \xmark      &  \cmark   
& \xmark & \xmark  & \cmark  &   \cmark &  \cmark & \xmark \\
\midrule
\rowcolor{gray!15}
\dataset    & Human      & 20,061       & \textbf{948}  & \textbf{230}        &  {\cmark }  & {\cmark } 
& {\cmark } & {\cmark } & {\cmark } & {\cmark } & {\cmark } & {\cmark } 
\\
\bottomrule
\end{tabular}
}
\centering
\end{table*}
Regarding EgoSchema, 71\% of its LLM-generated questions are centered on overall goals or primary actions, leading to limited diversity in question types (see Fig.~\ref{fig:q_type_comparison}(a)). In contrast, \dataset supports a broad range of reasoning skills, including temporal, causal, and hypothetical reasoning, through its carefully curated question design.
MovieChat-1k also shows limited diversity: 77\% of its questions are based on single frames (Fig.~\ref{fig:q_type_comparison}(b)). Moreover, all questions start with a fixed set of words such as ``do", ``who", ``which", ``when", ``how", or ``what", restricting the linguistic variety and limiting the inclusion of deeper reasoning questions, such as those starting with ``why" or ``what if". By comparison, \dataset includes a wider spectrum of question formats and reasoning levels.

VideoMME differs from our \dataset in three important ways. (1) \textit{Scale and usability}: \dataset contains 20,061 five-choice QA pairs with clear train/validation/test splits, enabling both model training and evaluation. In contrast, VideoMME offers only 2,700 four-choice QA pairs without a training split, limiting it to evaluation-only use. (2) \textit{Video length}: Our \dataset provides three hierarchical levels—single-scene (7.8 minutes), multi-scene (22 minutes), and full-scene (125.7 minutes)—which are significantly longer than VideoMME’s short ($<$2 min), medium (4–15 min), and long (30–60 min) clips. (3) \textit{Structural design}: VideoMME’s video categories are disjoint and based purely on duration, while \dataset constructs temporally coherent hierarchical levels from the same movie based on narrative structure.
Each movie in \dataset provides all three levels of granularity, enabling multi-scale reasoning within a unified narrative context.
This design provides a more challenging setting for long-form video understanding. Notably, while Gemini 1.5 Pro achieves 81.3\% accuracy on VideoMME, its performance drops to 71.5\% on \dataset, highlighting the increased difficulty and rigor of our \dataset. 
Regarding the Perception Test~\cite{patraucean2023perception} benchmark, its video clips average only 23 seconds in length and are not designed to support long-form video understanding.

\vspace{-2mm}
\subsection{Distractor Options Generation}
The role of distractor options in multiple-choice question-answering is crucial because they determine,  to some extent,  the difficulty of QA. 
Well-designed distractor options should closely resemble the correct answer in terms of plausibility, making it challenging for participants to select the correct option. 
One direct and labor-saving approach to generate distractor options is to use large language models (LLM) such as ChatGPT\cite{openai2022chatgpt}. However, we find that the generated candidates are unsatisfactory and have significant shortcuts. 
\subsubsection{Using LLMs for Distractor Options Generation}
\label{subsubsec:llm_distractor_generation}
\begin{figure}[t]
\begin{center}
\includegraphics[width=\linewidth]{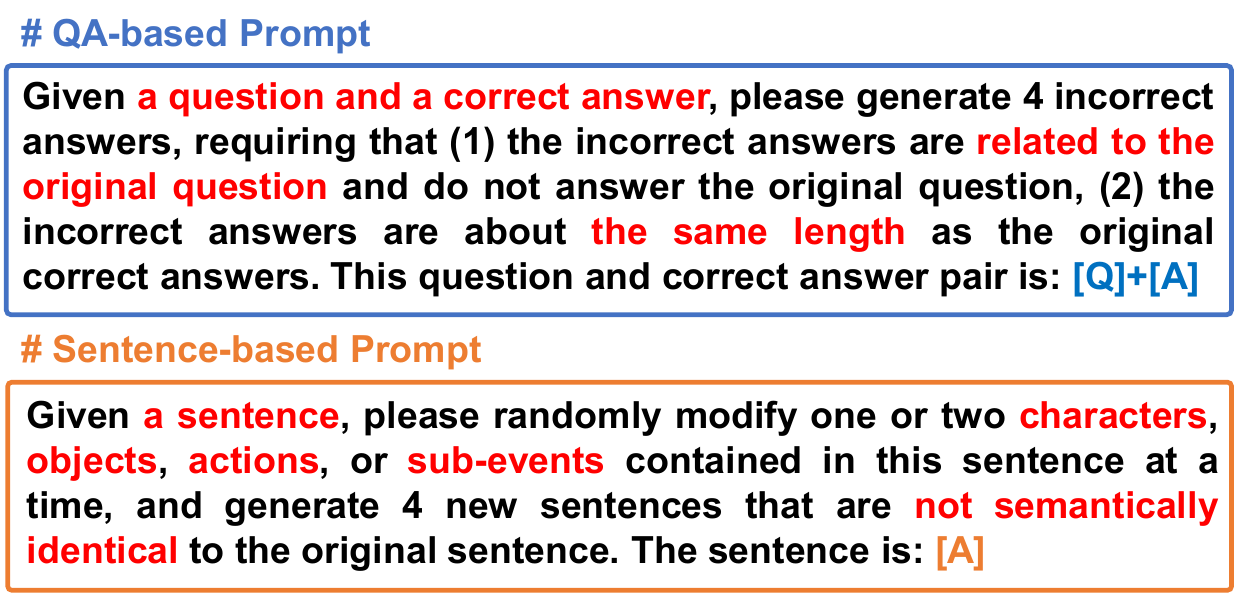}
\end{center}
\caption{Two types of prompts are used for generating distractor options by ChatGPT.
}
\label{fig:qa_prompt}
\end{figure}

\noindent\textbf{Prompts for Generating Distractor Options.}
Fig.~\ref{fig:qa_prompt} shows two types of prompts for generating distractor options using ChatGPT\cite{openai2022chatgpt}.
One type, given the question and correct answer, generates distractor options that cannot answer the question and are semantically different from the correct answer (QA-based). 
The other type,  given only the correct answer, generates distractor options that are different from the correct answer (Sentence-based).

\begin{figure}[t]
\begin{center}
\includegraphics[width=\linewidth]{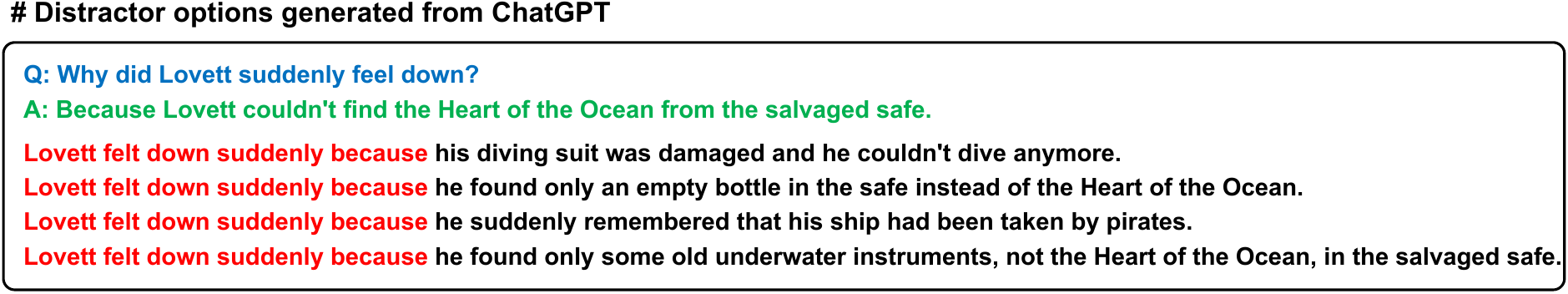}
\end{center}
\caption{Distractor options are generated by ChatGPT. Note that the distractor options are more likely to be longer and express similar sentence structures. Best viewed in color.
}
\label{fig:wrong_chatgpt}
\end{figure}

\noindent\textbf{Shortcuts in Generated Distractor Options.}
Fig.~\ref{fig:wrong_chatgpt} shows distractor options generated by ChatGPT. Compared to the correct option, these distractor options tend to have longer sentence lengths on one hand, which is due to the redundancy of ChatGPT compared to humans. On the other hand, these distractor options exhibit very similar wording in terms of sentence structure. 
These flaws become shortcuts for the model to select the correct answer, resulting in a high accuracy rate.
Therefore, we ultimately chose human annotators to design and generate the distractor options.

\subsubsection{Manual Creation of Distractor Options}
Due to the above drawbacks of using LLMs to generate distractor options, we opt to engage human annotators in the process of writing distractor options. 
We establish three guiding principles to avoid the bias from the distractor options generated by ChatGPT.
Firstly, we ensure that the length of the distractor options closely matches that of the correct options, to avoid overly long or short distractor bias.
Secondly, the semantic content of the distractor options are different from that of the correct answers. To achieve this, we employ SentenceTransformer~\cite{reimers2019sentence} to calculate the semantic differences and instruct annotators to verify and correct options with a semantic similarity exceeding 0.9.
Finally, for long answers, we mandate that annotators modify at least two distinct parts of the response to create a more robust set of distractor options.
We show in Section \ref{sec:exp;subsec:discussion} that human-corrected distractor options following these three principles are more challenging than those directly generated by ChatGPT\cite{openai2022chatgpt}.

Note that we do not ask human workers to generate distractors for our temporal perception QA, because the format of temporal perception answer is a time interval that can be generated according to rules. To generate difficult distractors for temporal perception QA, we randomly chose four equal-length, non-overlap intervals that are before or after the time interval of the correct answer. The equal-length design can avoid bias in the length distribution of time intervals.

\subsection{Quality Control}
To ensure high-quality annotations, we adopt a series of rigorous mechanisms, including pretesting the annotators before formal annotation, preparing informative reference slides and demonstrations, as well as cross-validating across annotators. 

\noindent\textbf{Collaborative Annotation.} We adopt a collaborative annotation approach, where each movie is annotated by at least two annotators. One annotator is responsible for asking questions, while the other provides the answers and corrects any mistakes. This dual-annotator system ensures a check-and-balance mechanism that reduces individual bias and increases the reliability of the annotations.

\noindent\textbf{Human Check.} We introduce a review stage to enhance the quality of the annotations further. Two independent reviewers who were not involved in the annotations meticulously examined the annotated QAs. These reviewers are incentivized with a cash bonus for identifying any mistakes, thus motivating them to be thorough and diligent in their review. This incentive structure not only promotes higher accuracy but also ensures that the reviewers are actively engaged in maintaining the highest standards of annotation quality.

\noindent\textbf{LLM Check.} To ensure that our questions are appropriately challenging and not overly simplistic, we implement a rigorous model-based check process. This involves using a LLM with fewer than 2 billion parameters, specifically InternLM~\cite{team2023internlm} with 1 billion parameters, to pre-screen the questions. The purpose of this step is to filter out any questions that the model can answer directly. By doing so, we aim to eliminate questions that might be too straightforward or obvious, ensuring that the remaining questions require a deeper understanding and more complex reasoning. This process helps maintain a high level of difficulty and relevance in our question set, making it more effective for evaluating the comprehension abilities of advanced models.

\section{Experiments} \label{sec:exp}
We use a 6:2:2 ratio split the 100 movies of \dataset dataset into train/val/test set. We ensure movies and their corresponding QA pairs appear in only one split.
This results in 11,923 QA pairs for training, 3,983 QA pairs for validation, and 4,155 QA pairs for testing. We use the \textcolor{blue}{multiple-choice QA} accuracy as the evaluation metric. We evaluate the performance of \dataset using traditional VideoQA methods, open-ended MLLM-based methods, and our proposed method with details below.

\vspace{-2mm}
\subsection{Baseline Methods}
\noindent\textbf{Traditional VideoQA methods.}
The traditional VideoQA methods treat multiple-choice QA task as a classification task and they output classification probability for every options. For traditional VideoQA methods, we choose to use recent works FrozenBiLM~\cite{yang2022zero}, BLIP-2~\cite{li2023blip}, and SEVILA as baselines.
FrozenBiLM treats the options individually and outputs the yes/no probability for each option, where the final choice is determined by the option with the highest yes probability. BLIP-2 utilizes all choice options as input. The output probabilities on the tokens A/B/C/D/E of the whole word logits are held out to get the predicted probabilities on each option. SEVILA first selects keyframes using a localizer, then follows the procedure of BLIP-2 to generate the answer choices. 
\begin{figure*}[t]
\begin{center}
\includegraphics[width=0.95\linewidth]{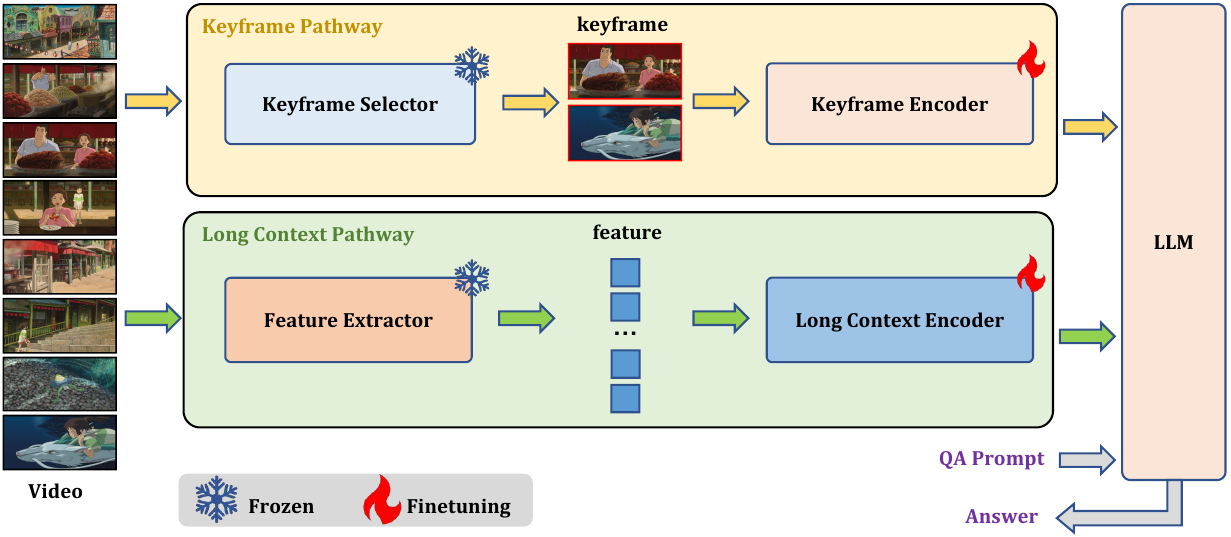}
\end{center}
\caption{Our \method framework includes a keyframe pathway and a long context pathway. A similar process is used for the subtitle modality and thus omitted for simplicity.}
\label{fig:method_ours}
\end{figure*}

\noindent\textbf{Dual-Pathway Modeling (\method).}
In addition, to effectively address the challenges of both long video length and clue length in long-form video QA, we propose a direction called Dual-Pathway Modeling (\method). From a human's perspective, when answering a question about a long video, it is essential to consider both the key clues related to the question and the contextual elements of the movie. Following this intuition, we take an initial step and design a \method framework for long-form video QA.

The DPM contains two branches: keyframe pathway and long context pathway. As shown in Fig.~\ref{fig:method_ours}, the keyframe pathway is intended for selecting the most important clues. It includes a keyframe selector which is an improved version of the Localizer~\cite{yu2023self}  with both frames and their subtitles as selection criteria. A keyframe encoder is then used to encode the keyframes. As a complementary, the long context pathway models the overall contextual information in the long videos. It starts with extracting video features using a pretrained feature extractor. These video features are then processed by a long-context encoder, which follows the architecture of the Q-Former model from~\cite{li2023blip}. The long-context encoder summarizes the features from the entire video and compresses them into fixed-length tokens.
Finally, we concatenate the output of the two pathways and the QA prompt as the input to the LLM to generate an answer.

\noindent\textbf{MLLM-based methods.}
MLLM-based methods demonstrate their ability as general models by generating free-form text for VideoQA. Therefore, their predictions may not be included in the five candidate options.
The final choice is selected by matching the selections to one of the options by rule-based~\cite{li2024mvbench} or similarity-based matching~\cite{maaz2023video}.
The MLLM methods consist of open-source and closed-source MLLM methods. The open-source MLLM methods include VideoChat ~\cite{li2023videochat}, Video-ChatGPT~\cite{maaz2023video}, mPLUG-Owl~\cite{ye2023mplug}, Otter~\cite{li2023otter}, VideoChat2~\cite{li2024mvbench}, Qwen2-VL~\cite{wang2024qwen2}, LongVA~\cite{zhang2024long}, LLaVA‑Video~\cite{zhang2024video} and LLaVA‑OV~\cite{li2024llava}. We directly apply the built-in solutions for multiple-choice QA for these methods.
The closed-source MLLM methods include 
Gemini-1.5 Pro~\cite{reid2024gemini} and GPT-4o~\cite{gpt4o}.
For these methods, we use the same matching mechanism as VideoChat2~\cite{li2024mvbench} to generate the final output choice. We found that earlier multimodal LLMs often struggled with this requirement, whereas recent models perform flawlessly.  
For instance, both VideoChat2~\cite{li2024mvbench} and Qwen2-VL~\cite{wang2024qwen2} produce valid choices 100\% of the time, while VideoChat~\cite{li2023videochat} does so only 71\% of the time.

\noindent\textbf{Implement Datails$\quad$}
In our \method, we uniformly sample 512 frames. For the keyframe pathway, we select top 32 frame and subtitle pairs. 
For the long context pathway, the learnable query embeddings for long frames and long subtitles are both 32.
All experiments are conducted on 8 A100 GPUs. Due to the large computation of the BLIP-2, we can only use a batch size of 1. Regarding other hyperparameters, we refer to the settings in~\cite{yu2023self}.
Table.~\ref{tab:hypers} shows the fine-tuning hyperparameters of our \method.
As for other VideoQA methods and all MLLM methods, their experimental settings follow the original paper, except that we input additional subtitles for multimodal reasoning.
For GPT-4o~\cite{gpt4o}, we uniformly sample 384 frames as visual inputs. 
For Gemini-1.5 Pro~\cite{reid2024gemini}, we uniformly sample 1000 frames.

\begin{table}[ht]
\caption{DPM fine-tuning hyperparameters.}
\label{tab:hypers}
\centering
\scalebox{0.77}{
\begin{tabular}{ccccc }
\toprule
All Frames & Keyframes & Learning Rates & Epochs &  Gradient Accumulation Step
 \\
\midrule
 512  & 32 & 3e-5 & 10 & 1\\
\bottomrule
\end{tabular}
}
\end{table}

\begin{table*}[t]
\caption{The multiple-choice QA accuracy of traditional VideoQA methods on different scenes and QA types in \dataset.}
\label{tab:mc_compare_q_scene_type_1}
\centering
\scalebox{0.825}{
\begin{tabular}{cccccccccccccc}
\toprule
Setting & Method & \makecell{LLM \\ Params} & \makecell{Sampled\\Frames}
  & Single & Multi & Full
  & Synopsis & Temporal & Spatial & Causal & Hypothetical & Knowledge & Overall \\
\midrule
\multirowcell{3}{Zero-shot}
  & FrozenBiLM \cite{yang2022zero}
    & 0.9B & 64
    & 25.9 & 22.1 & 21.2
    & 27.9 & 20.7 & 28.0 & 21.5 & 21.3 & 30.2 & 24.8 \\
  & BLIP-2 \cite{li2023blip}
    & 3B & 64
    & 28.2 & 26.0 & 25.0
    & 28.1 & 23.1 & 30.8 & 25.8 & 27.1 & 38.7 & 27.6 \\
  & SEVILA \cite{yu2023self}
    & 3B & 512
    & \bf 28.9  & \bf 26.3 & \bf 25.1
    & \bf 28.4  & \bf 23.4  & \bf 34.9  & \bf 26.0 & \bf 27.3  & \bf 39.3  & \bf 28.2 \\
\midrule
\multirowcell{4}{Finetuning}
  & FrozenBiLM \cite{yang2022zero}
    & 0.9B & 64
    & 32.1 & 30.6 & 25.7
    & 34.7 & 22.2 & 32.6 & 34.7 & 32.5 & 40.1 & 31.5 \\
  & BLIP-2 \cite{li2023blip}
    & 3B & 64
    & 35.3 & 32.1 & 29.1
    & 36.4 & 23.9 & 35.6 & 40.5 & 36.5 & 45.1 & 34.3 \\
  & SEVILA \cite{yu2023self}
    & 3B & 512
    & 36.1 & 32.5 & 29.4
    & 37.4 & 24.3 & 35.9 & 41.6 & 36.9 & 45.7 & 35.0 \\
 \rowcolor{gray!15}
  & \method
    & 3B & 512
    & \bf 37.6 & \bf 35.7 & \bf 31.3
    & \bf 40.8 & \bf 24.5 & \bf 37.6 & \bf 43.2 & \bf 39.8 & \bf 46.0 & \bf 36.9 \\
\bottomrule
\end{tabular}
}
\end{table*}
\subsection{Results of Traditional VideoQA Methods}
The results of traditional VideoQA methods are reported in Table \ref{tab:mc_compare_q_scene_type_1} for both zero-shot and finetuning settings, considering different scenes and QA types. 
Firstly, SEVILA~\cite{yu2023self} demonstrates the strongest zero-shot performance among all baselines, achieving an overall accuracy of 28.2\%, while our proposed method (\method) achieves the best finetuning results, with an accuracy of 36.9\%. 
Across all methods, the zero-shot accuracies of all methods range from 25.1\% to 28.9\%, and the finetuning accuracies range from  31.3\% to 37.6\%, indicating a moderate improvement when finetuning is applied. 
Since the chance-level accuracy is 20\%, the results reveal that it remains very challenging for most traditional VideoQA methods to understand long-form videos like movies.
Secondly, a deeper analysis reveals that performance consistently declines as the length of scenes increases, emphasizing the challenges associated with answering questions that require understanding longer video contexts. This trend underscores the difficulty of processing long-form clues and validates the quality of the dataset annotations, which intentionally include extended video and clue lengths. 
Moreover, Table \ref{tab:mc_compare_q_scene_type_1} also presents the results for different QA types. Notably, the temporal perception QA type presents the most significant challenge. This finding underscores the limitations of existing methods in effectively addressing this specific aspect of temporal perception QA~\cite{ren2023timechat}, highlighting the need for further advancements in modeling temporal relationships.

\begin{table*}[t]
  \centering
  \caption{The zero-shot multiple-choice QA accuracy of MLLM methods on different scenes and QA types in our \dataset.}
  \label{tab:open_ended_llm_method_rebuttal}
  \begin{adjustbox}{max width=1\linewidth}
    \begin{tabular}{
      lcccccccccccc
    }
      \toprule
      Models
      & \makecell{LLM\\Params}
      & \makecell{Sampled\\Frames}
      & Single
      & Multi
      & Full
      & Synopsis
      & Temporal
      & Spatial
      & Causal
      & Hypothetical
      & Knowledge
      & Overall \\
      \midrule

      \multicolumn{13}{c}{\itshape Open-source Video MLLMs} \\
      \midrule
      Mplug-Owl~\cite{mplugowl}      & 7B  & 16  & 25.2 & 23.5 & 22.1 & 25.1 & 19.9 & 25.3 & 21.9 & 23.5 & 27.5 & 24.7 \\
      Otter~\cite{li2023otter}       & 7B  & 16  & 23.1 & 22.1 & 21.3 & 22.6 & 20.7 & 19.6 & 26.1 & 24.2 & 21.8 & 22.8 \\
      VideoChatGPT~\cite{maaz2023video} & 7B & 16  & 23.4 & 22.7 & 22.3 & 23.8 & 20.2 & 22.1 & 22.1 & 21.4 & 24.1 & 23.2 \\
      VideoChat~\cite{li2023videochat}  & 7B & 16  & 24.9 & 24.1 & 21.6 & 25.2 & 20.1 & 26.7 & 25.8 & 22.9 & 26.0 & 24.6 \\
      VideoChat2~\cite{li2024mvbench}   & 7B & 16  & 27.2 & 26.3 & 24.2 & 30.0 & 21.7 & 29.3 & 26.7 & 25.8 & 29.6 & 26.9 \\
      LongVA~\cite{zhang2024long}       & 7B & 64  & 39.4 & 35.0 & 25.6 & 45.6 & 28.8 & 37.8 & 34.6 & 32.4 & 46.2 & 37.9 \\
      LLaVA-Video~\cite{zhang2024video}  & 7B & 64  & 42.0 & 34.5 & 24.5 & 49.7 & 27.7 & 47.8 & 35.2 & 29.6 & 45.0 & 39.6 \\
      LLaVA-Video~\cite{zhang2024video}  & 72B & 64 & 51.0 & 43.5 & 26.2 & 58.3 & 38.1 & 49.6 & 43.2 & 38.2 & {\bfseries 58.0} & 48.3 \\
      Qwen2-VL~\cite{wang2024qwen2}     & 7B  & 648 & 42.3 & 40.0 & 29.4 & 49.4 & 28.9 & 47.8 & 39.8 & 36.1 & 47.5 & 41.3 \\
      Qwen2-VL~\cite{wang2024qwen2}     & 72B & 648 & 49.7 & {\bfseries 45.3} & {\bfseries 41.2} & 58.8 & 38.5 & {\bfseries 50.6} & 42.6 & 37.4 & 56.5 & 48.4 \\
      LLaVA-OV~\cite{li2024llava}     & 7B & 64  & 42.6 & 36.0 & 26.5 & 50.3 & 26.8 & 47.8 & 36.7 & 35.3 & 45.6 & 40.4 \\
      LLaVA-OV~\cite{li2024llava}     & 72B & 64 & {\bfseries 52.0} & 44.9 & 29.1 & {\bfseries 59.0} & {\bfseries 41.2} & 50.4 & {\bfseries 43.8} & {\bfseries 38.4} & 57.4 & {\bfseries 49.4} \\
      \midrule
      \multicolumn{13}{c}{\itshape Closed-source Video MLLMs} \\
      \midrule
      GPT-4o~\cite{gpt4o}               & –   & 384 & 68.8 & 63.9 & 59.8 & 71.4 & 62.5 & 67.8 & 70.5 & 66.4 & 61.3 & 67.3 \\
      Gemini-1.5 Pro~\cite{reid2024gemini} & – & 1000 & {\bfseries 73.2} & {\bfseries 67.7} & {\bfseries 63.1} & {\bfseries 75.4} & {\bfseries 65.6} & {\bfseries 71.4} & {\bfseries 72.2} & {\bfseries 70.8} & {\bfseries 66.0} & {\bfseries 71.5} \\
      \bottomrule
    \end{tabular}
  \end{adjustbox}
\end{table*}

\subsection{Results of MLLM Methods}
We report the zero-shot performance of state-of-the-art open-source and closed-source MLLM methods in Table~\ref{tab:open_ended_llm_method_rebuttal}. 
Upon analyzing the results, we observe that LLaVA-OV 72B~\cite{li2024llava} stands out with an overall score of 49.4\%, significantly outperforming other open-source models. We attribute this strong performance to its high-quality pretraining data. In particular, LLaVA-OV is trained on a diverse corpus of 1.6 million samples annotated by GPT-4V/o and Gemini, encompassing single-image, multi-image, and video-based inputs.
Moreover, we find that the performance of all MLLM models deteriorates as the length of the video segments increases. This trend suggests that longer video segments pose greater challenges, aligning with similar findings in the evaluation of traditional VideoQA methods. Among all evaluated models, the closed-source MLLM Gemini 1.5 Pro~\cite{reid2024gemini} achieves the best overall performance, with a score of 71.5\% on the \dataset dataset.
Interestingly, closed-source MLLMs outperform all open-source counterparts. A possible explanation for the superior performance of leading closed-source models is their training on substantially larger corpora, which may endow them with stronger long-form video understanding capabilities.

\subsection{Discussions} \label{sec:exp;subsec:discussion}

\begin{table*}[t!]
\caption{
Human performance with different modality input. Modality inputs Q, V, and S represent question, video, and subtitle, respectively.
}
\label{tab:human_different_modality}
\centering
\scalebox{0.9}{
\begin{tabular}{cccccccccccc}
\toprule
Method & Modality input &  Single& Multi & Full & Synopsis & Temporal & Spatial & Causal & Hypothetical  & Knowledge & Overall  \\
\midrule
\multirowcell{4}{Human} 
& Q & 40.5 & 39.8 & 37.2 & 46.9 & 20.3 & 45.1 & 43.6 & 42.8 & 60.5  & 40.2  \\
& V + Q & 74.9 & 71.6 & 64.2  & 71.3 & 76.2 & 75.1 & 73.5 & 71.5 & 78.3 & 73.7  \\
& S + Q & 66.3 & 64.7 & 57.1 & 72.8 & 45.2 & 68.5 & 72.1 & 68.3 & 83.8 & 65.6 \\
& V + S + Q & \bf 92.1 & \bf 89.5 & \bf 86.3 & \bf 91.7 & \bf 90.3 &  \bf91.5 & \bf 92.4 & \bf 91.3 &  \bf 91.5 & \bf 91.3 \\
\bottomrule
\end{tabular}
}
\end{table*}

\noindent\textbf{Human Performance.
}  
To assess the human upper-bound on our \dataset and understand the effect of each modality in the video question-answering process, we asked human workers to answer questions from our \dataset using different modality inputs. 
The results, summarized in Table \ref{tab:human_different_modality}, demonstrate that human performance surpasses that of the state-of-the-art GPT-4o model reported in Table \ref{tab:open_ended_llm_method_rebuttal}. This finding highlights the difficulty of our \dataset and emphasizes the potential for further improvement in this area. Furthermore, we observed that even for humans, the performance declines as the number of scenes increases. This suggests that the task becomes more challenging when dealing with longer videos and more complex scenes, aligning with the trends observed in the evaluation of traditional VideoQA and MLLM methods. Analyzing different question types, we found that the visual input greatly helps humans in answering synopsis, temporal, and spatial questions. For example, the combination of visual input, subtitle input, and question input (V+S+Q) outperforms using only subtitle input and question input (S+Q) by over 40\% for temporal QA. This indicates that our QA annotation indeed depends on visual information, distinguishing our dataset from others like MovieQA\cite{tapaswi2016movieqa} and EgoSchema\cite{mangalam2023egoschema}.
On the other hand, the textual modality primarily aids humans in processing synopsis and knowledge questions. For example, for the knowledge questions, using subtitles and questions (S+Q) can outperform video and question (V+Q) input. This highlights the complementary nature of visual and textual modalities in addressing different types of questions and emphasizes the importance of multi-modal approaches in our \dataset benchmark. 
It is also important to note that human participants, equipped with a wealth of general knowledge, can answer 40\% of questions correctly with only the question. A phenomenon observed across datasets like TVQA\cite{lei2018tvqa}, where question-only accuracy is also above random chance (around 32\%), indicating that some questions rely on common sense or widely known information rather than video-specific content.

\begin{table*}[t]
\caption{Ablation study of different distractor option generation strategies for multiple-choice QA.}
\label{tab:mc_compare_q_wrong_answer_1}
\centering
\scalebox{1}{
\begin{tabular}{cccccc}
\toprule
Method & Distractor Option Type & Single-Scene & Multi-Scene & Full-Scene & Overall  \\
\midrule
\multirow{3}{*}{FrozenBiLM \cite{yang2022zero}} 
& QA-based  & 79.2 & 80.9 & 79.4  & 79.6\\
& Sentence-based  & 55.3 & 51.2 & 52.3  & 54.2\\
& Manual Annotation  & 32.1  & 30.6 &  25.7 &  31.5 \\
\bottomrule
\end{tabular}
}
\end{table*}
\noindent\textbf{Human-Written vs ChatGPT.}
We initially attempted to generate distractor options using ChatGPT\cite{openai2022chatgpt}. We conducted preliminary experiments using the FrozenBiLM method~\cite{yang2022zero} with its original hyperparameters.
We tried two types of prompts for generating distractor options using ChatGPT. 
One is given the question and correct answer, it generates distractor options that cannot answer the question and are semantically different from the correct answer (QA-based). 
Another is given only the correct answer, it generates distractor options that are different from the correct answer (Sentence-based).
The detailed prompt can be found in Section \ref{subsubsec:llm_distractor_generation}.
From Table.~\ref{tab:mc_compare_q_wrong_answer_1}, we can see that both ChatGPT-prompt-generated QA datasets have very high QA accuracy, and our manual distractor options lead to a much lower accuracy. This indicates there may exist significant QA shortcuts for ChatGPT-generated distractor options. We found that many distractor options generated by ChatGPT are consistently either too short or too long compared to the correct answer, and sometimes use different words but convey a similar semantic meaning as the correct answers.
An example can be seen in Fig. \ref{fig:wrong_chatgpt}.
Generating reliable multiple-choice QA based on ChatGPT may be a good avenue for future work.

\noindent\textbf{Frame Input Variability for GPT-4o and Gemini-1.5 Pro.}
\begin{table}
\caption{The performance of GPT-4o and Gemini-Pro with different numbers of frame inputs.}
\label{tab:gpt_different_frame}
\centering
\scalebox{0.88}{
\begin{tabular}{cccccc}
\toprule
Method & Sampled Frames & Single & Multi & Full & Overall \\ \hline
\multirow{3}{*}{\begin{tabular}[c]{@{}c@{}}GPT-4o \cite{gpt4o} \\ (V+S+Q)\end{tabular}} & 10 & 55.7 & 45.4 & 43.8 & 52.8  \\ 
 & 100 & 65.6 & 54.7 & 51.8 & 62.5  \\  
 & 384 & \bf 68.8 & \bf 63.9 & \bf 59.8 & \bf 67.3 \\ 
 \hline
\multirow{3}{*}{\begin{tabular}[c]{@{}c@{}}Gemini-1.5 Pro \cite{reid2024gemini} \\ (V+S+Q)\end{tabular}} & 10 & 58.7 & 53.3 & 49.6 & 57.2 \\ 
 & 100 & 66.7 & 60.9 & 57.9 & 65.0  \\  
 & 1000 & \bf 73.2 & \bf 67.7 & \bf 63.1 & \bf 71.5 \\ 
\bottomrule
\end{tabular}}
\end{table}
Table.~\ref{tab:gpt_different_frame} shows the performance of GPT-4o~\cite{gpt4o} and Gemini-Pro~\cite{reid2024gemini} with different numbers of frame inputs. We use 384 frames for GPT-4o due to its 128k context size.
Based on the results, we observe that the performance of both GPT-4o and Gemini-Pro improves with an increased number of frame inputs. For GPT-4o, using 384 frames yields the highest overall score of 67.3\%, indicating a substantial improvement over using just 10 or 100 frames. This can be attributed to the model's ability to effectively utilize its context, which allows for more comprehensive visual analysis.
Similarly, Gemini-1.5 Pro shows a significant enhancement in performance with 1000 frames, achieving an overall score of 71.5\%. This suggests that Gemini-Pro benefits from a larger number of visual inputs, likely due to its advanced processing capabilities which can handle extensive visual data efficiently.
In summary, both models demonstrate that increasing the number of frame inputs can lead to improved performance across different tasks, highlighting the importance of selecting an appropriate number of frames to leverage the full potential of these models.

\begin{table}
    \centering
    \captionsetup{width=\linewidth} 
    \caption{
        Ablation study of our \method. The variants \method-K and \method-C indicate that only the keyframe pathway or long context pathway is used for \method. DPM (N) means N frames are inputs to DPM, and the default N is 512.
    }
    \label{tab:dpm_ablation}
    \scalebox{0.9}{
    \begin{tabular}{ccccc}
        \toprule
        Method  & Single & Multi & Full & Overall  \\
        \midrule
        \method-C    &  36.9  & 34.1  & 30.3 & 36.0 \\
        \method-K &  37.2 &  34.6 &  29.9 &  36.3  \\ 
        \method (128)   & 37.1 &  34.4  & 30.4 & 36.2  \\
        \method (256)   & 37.4 &  35.1  & 30.9 & 36.6  \\
        \method (512)   &  \bf37.6   &  \bf 35.7 &  \bf  31.3  & \bf 36.9  \\
        \bottomrule
    \end{tabular}}
\end{table}

\begin{table}[t]
  \centering
  \caption{
  Multiple-choice accuracy on \dataset when DPM is applied on different MLLM backbones.
  }
  \label{tab:dpm_different_backbone}
  \begin{adjustbox}{max width=\linewidth}
    \begin{tabular}{
      lcccccc
    }
      \toprule
      Method
      & \makecell{LLM\\Params}
      & \makecell{Sampled\\Frames}
      & Single
      & Multi
      & Full
      & Overall \\
      \midrule
      SEVILA~\cite{yu2023self}               & 3B & 512 & 36.1 & 32.5 & 29.4 & 35.0 \\
      DPM (SEVILA)                          & 3B & 512 & {\bfseries 37.6} & {\bfseries 35.7} & {\bfseries 31.3} & {\bfseries 36.9} \\
      \midrule
      LongVA~\cite{zhang2024video}          & 7B & 64  & 46.6 & 40.3 & 29.0 & 44.4 \\
      DPM (LongVA)                          & 7B & 512 & {\bfseries 48.4} & {\bfseries 43.5} & {\bfseries 31.4} & {\bfseries 46.6} \\
      \midrule
      LLaVA-OV~\cite{li2024llava}           & 7B & 64  & 48.9 & 41.6 & 30.3 & 46.5 \\
      DPM (LLaVA-OV)                        & 7B & 512 & {\bfseries 50.5} & {\bfseries 44.1} & {\bfseries 32.4} & {\bfseries 48.3} \\
      \bottomrule
    \end{tabular}
  \end{adjustbox}
\end{table}
\noindent\textbf{Variants of \method.}
To analyse the effectiveness of different components in our \method, we conduct ablation experiments in Table \ref{tab:dpm_ablation}.
We evaluate the performance of \method-K, the variant of \method with only the keyframe pathway, and \method-C, \method with only the long-context pathway. Surprisingly, \method-K achieved the second-best performance. This suggests that the keyframe branch alone can capture important information for many video QAs. The combination of two pathways can lead to performance improvement, which validates our motivation that two pathways can provide complementary information. Additionally, we study the impact of varying the number of input frames on performance. The results indicate that incorporating a larger temporal context can enhance the performance of our \method. It is also possible to extend the number of input frames beyond 512, however at the cost of speed and efficiency. We leave the exploration as our future work. 

We further demonstrate the generality and plug-and-play nature of our DPM module by integrating it into three representative VideoQA backbones: SEVILA, LongVA, and LLaVA-OV. The results, summarized in Table~\ref{tab:dpm_different_backbone}, show that in all cases, the DPM-augmented variants consistently outperform their base counterparts. Specifically, DPM~(SEVILA) achieves a 1.9\% improvement over SEVILA, DPM~(LongVA) yields a 2.2\% gain over LongVA, and DPM~(LLaVA-OV) delivers a 1.8\% boost over LLaVA-OV. These findings indicate that DPM is backbone-agnostic and can be seamlessly integrated into a wide range of long-form VideoQA frameworks to enhance their performance.

\begin{figure}
    \centering
    \includegraphics[width=\linewidth]{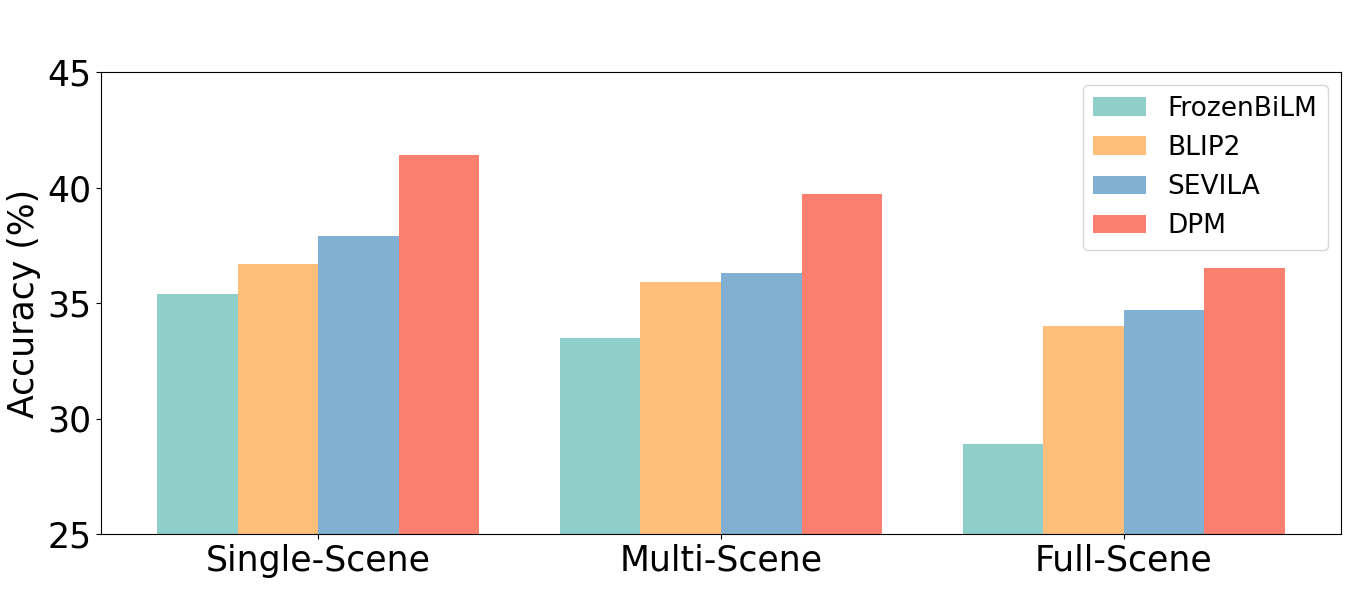} 
    \caption{Information synopsis QA performance of traditional VideoQA methods in different scenes. Best viewed in color.}
    \label{fig:info-scene}
\end{figure}
\noindent\textbf{Impact on The Length of Scenes.}
For a more direct comparison, we present the performance of models on the same QA type across different scenes. Fig. \ref{fig:info-scene} shows the QA accuracy of 4 methods for the Information Synopsis questions on single-scene, multi-scene, and full-scene. Clearly, as the temporal length of the scenes increases, the difficulty of the questions also becomes higher, which is reflected by the unanimous performance decrease in all models. This finding aligns with our expectations and is consistent with previous observations in video question answering, showing the significant room for future work to improve upon.

\section{Conclusions}
We present \dataset, a long-form videoQA dataset, and construct a benchmark to assess the versatile cognitive capabilities of multimodal systems.
Our \dataset features significantly longer video and clue length compared to the existing VideoQA datasets.
To move towards human-level understanding, QAs in our \dataset are manually labeled from the perspective of the interconnected abilities required by moviegoers to understand movie content. 
Our experiments suggest that even the most advanced models achieve unsatisfactory results. Our established \method has shown some improvements, but there is still ample scope for enhancement on our challenging \dataset dataset.
We believe that our \dataset will have a significant impact on the advancement and assessment of forthcoming long-form video understanding models.

\noindent\textbf{Limitations.}
There are a thousand Hamlets in a thousand people's eyes. Despite conducting multiple rounds of cross-validation, the understanding of the essence of a movie may vary among individuals. As a result, for some challenging abstract questions, the answers may not be entirely accurate.
Further, it is important to acknowledge that human curation is an imperfect process. Despite implementing multiple rounds of checks to minimize false positives, it is inevitable that the collected \dataset may include some mislabeled or improperly formatted QA pairs. Moreover, the dataset's scope is relatively limited as it predominantly relies on movies, which may restrict its applicability to broader video domains.

\noindent\textbf{Broader Impacts.}
Our \dataset has both positive and negative impacts. For the positive impacts, our \dataset can significantly contribute to the advancement of long-form video understanding and provides a rich resource for training models to understand complex video content and answer questions accurately. Meanwhile, the development of more sophisticated video understanding models can have numerous real-world applications, such as accessibility services for visually impaired individuals. By using movies as the source of the dataset, this work can also promote a deeper understanding and appreciation of film content, storytelling techniques, and cinematic history.
For the negative impacts, it’s crucial to address privacy concerns and the risk of deepfakes to ensure the responsible use of these technologies.

\noindent\textbf{Future Work.}
In the future, we plan to further enhance the quality of the dataset by refining its annotations and incorporating a broader variety of question types. Additionally, we aim to propose a more powerful framework for long-video understanding, capable of addressing the challenges posed by complex temporal and semantic structures. Furthermore, we intend to expand the scope of data sources beyond movies to include a wider range of video content, enabling more comprehensive and diverse video analysis.

\noindent\textbf{Data Availability Statements.}
Our annotation files are included in the supplementary materials.

\begin{acknowledgements}
This work is supported by National Key R\&D Program of China (2022ZD0160101), Industry Collaboration Projects Grant, Shanghai Committee of Science and Technology, China (Grant No. 22YF1461500), Jiangsu Frontier Technology Research and Development Program (No. BF2024076), and JSPS KAKENHI Grant Number JP25K24384.
\end{acknowledgements}

\appendix
\section*{Appendix}
\section{The 100 movies used in our \dataset} \label{appx:100movies} 
Table~\ref{tab:movie_table} shows the names, release years, and themes of the 100 movies used in our \dataset{} dataset. The movies cover a wide variety of themes, demonstrating significant differences in narrative focus. This diversity makes our dataset a valuable resource for long-form video understanding.
We order them by the IMDb score. 

\begin{figure*}
    \centering
    \includegraphics[width=\linewidth]{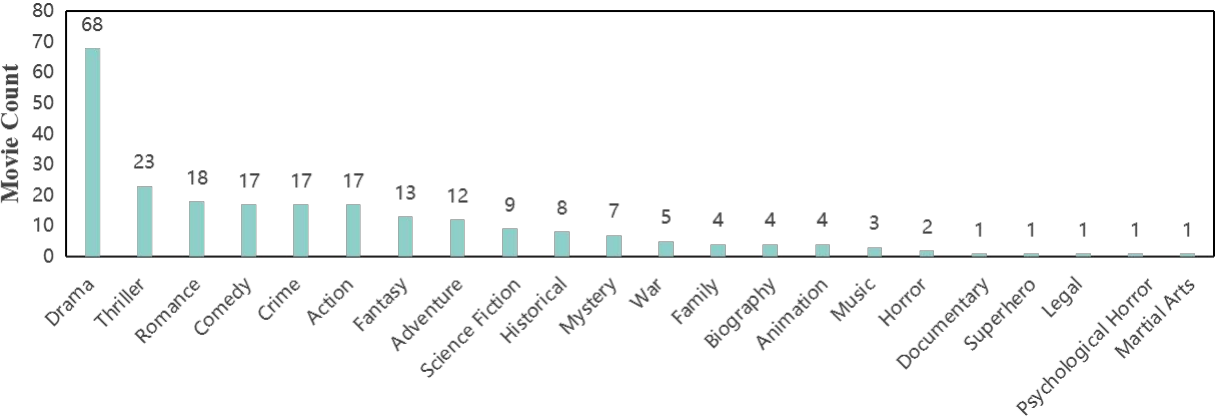}
    \caption{\textcolor{blue}{Genre distribution of the 100 movies in the \dataset dataset. Note that each movie may belong to multiple genres.}}
    \label{fig:genre}
\end{figure*}

Fig.~\ref{fig:genre} shows the genre distribution of the 100 movies in the \dataset{} dataset. Note that each movie may belong to multiple genres. The dataset exhibits a high degree of genre diversity, with drama being the most prevalent, followed by thriller, romance, comedy, crime, and action. A wide range of other genres—including fantasy, adventure, science fiction, historical, and mystery—are also well covered. Additionally, the presence of niche genres such as documentary, psychological horror, and martial arts further illustrates the thematic richness of the collection. This diversity provides a solid foundation for developing and evaluating models on complex long-form video understanding tasks across varied narrative and stylistic domains.

\begin{figure*}
    \centering
    \includegraphics[width=\linewidth]{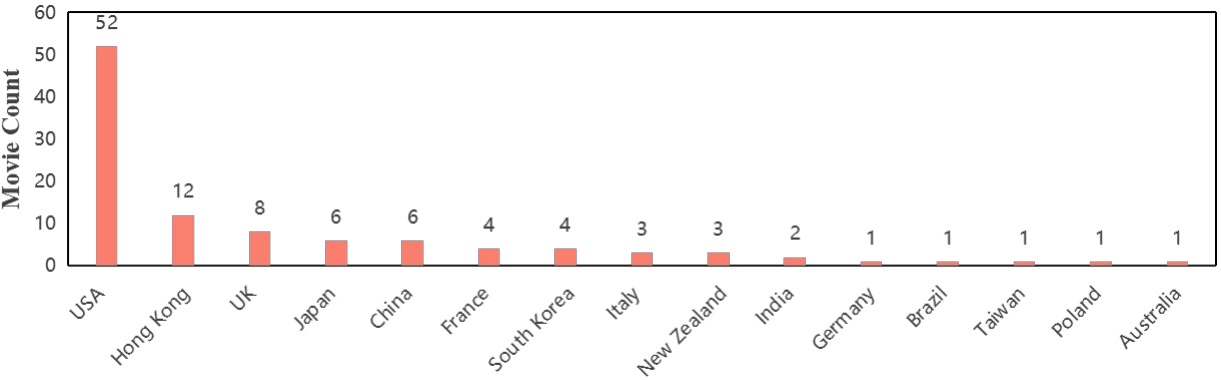}
    \caption{\textcolor{blue}{Country and region distribution of the 100 movies in the \dataset{} dataset. Note that some movies are co-produced by multiple countries or regions.}}
    \label{fig:country}
\end{figure*}

Fig.~\ref{fig:country} shows the country and region distribution of the 100 movies in the \dataset{} dataset. Note that some movies are co-produced by multiple countries or regions. The majority of the films originate from the United States, which accounts for over half of the dataset. Hong Kong, the United Kingdom, Japan, China, and France also contribute a notable number of titles. In addition, the dataset includes films from a diverse range of regions such as South Korea, Italy, New Zealand, India, and several others. This international distribution highlights the cultural diversity present in \dataset{}, making it well-suited for research on globally representative long-form video understanding.

\section{Examples of Various QA Types} \label{appx:datasets_examples}
\begin{figure*}
\begin{center}
\includegraphics[width=0.9\linewidth]{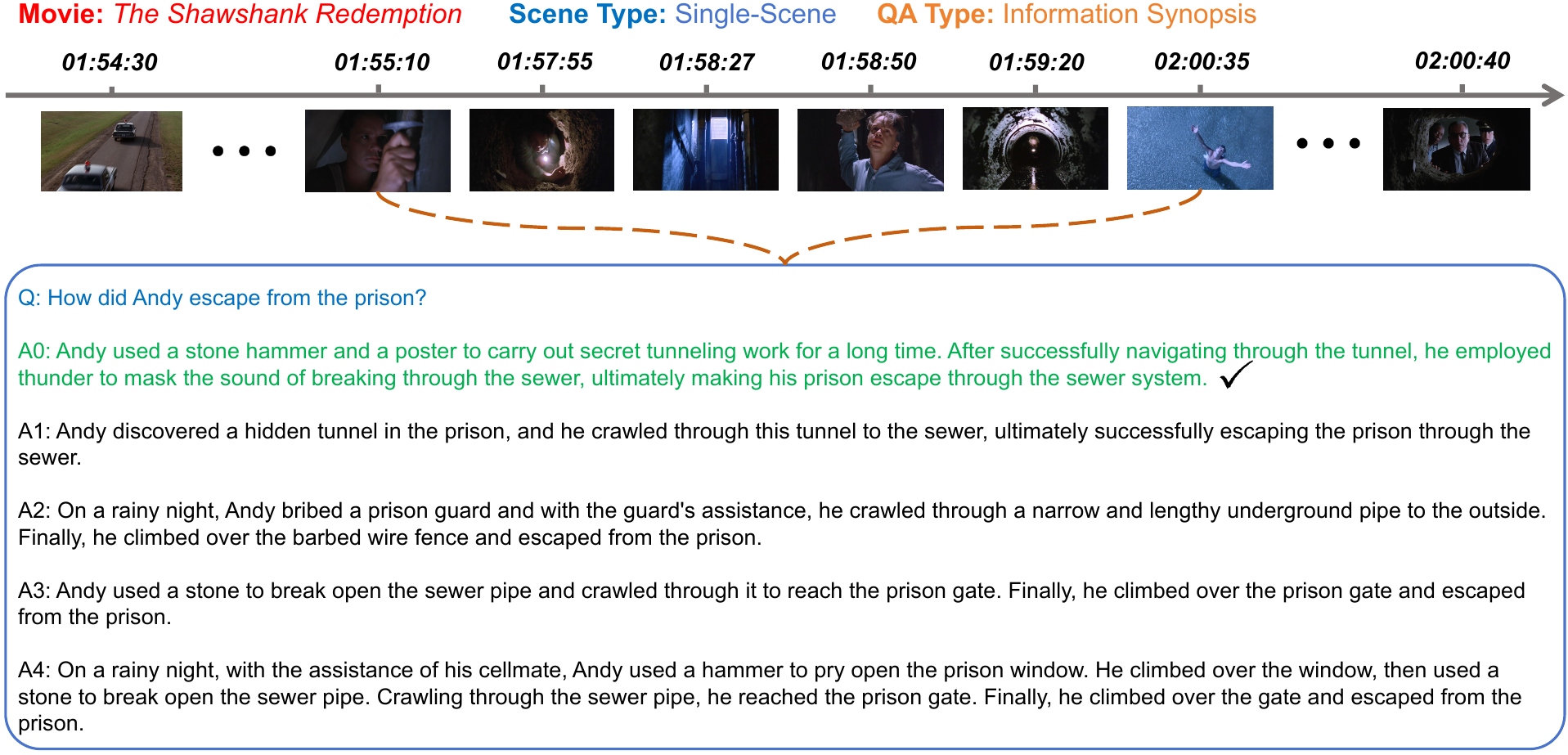}
\end{center}
\vspace{-5mm}
\caption{An Information Synopsis QA example from the \dataset dataset, as well as the prediction from our \method. The ground truth answer is highlighted in green, and the model prediction is indicated by \ding{51}. Best viewed in color.}
\vspace{-2mm}
\label{fig:qa_example1}
\end{figure*}

\begin{figure*}
\begin{center}
\includegraphics[width=0.9\linewidth]{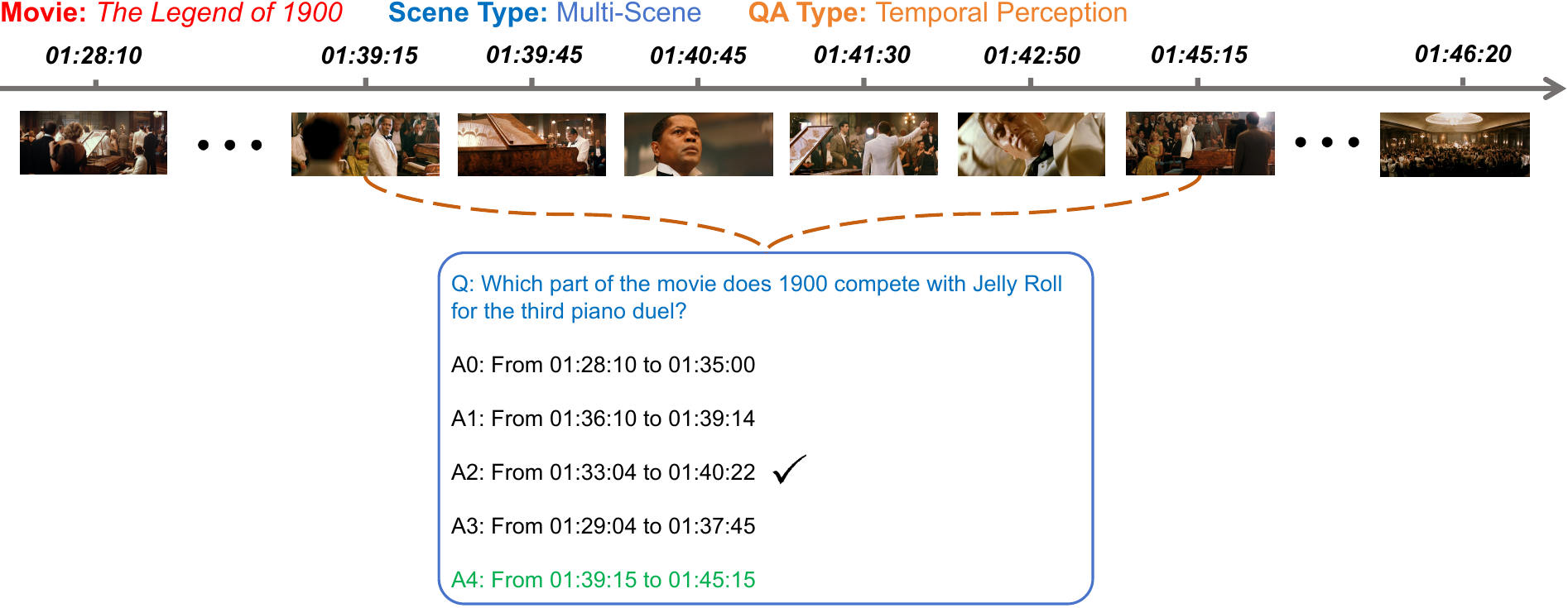}
\end{center}
\vspace{-5mm}
\caption{A Temporal Perception QA example from the \dataset dataset, as well as the prediction from our \method. The ground truth answer is highlighted in green, and the model prediction is indicated by \ding{51}. Best viewed in color.}
\vspace{-2mm}
\label{fig:qa_example2}
\end{figure*}

\begin{figure*}
\begin{center}
\includegraphics[width=0.9\linewidth]{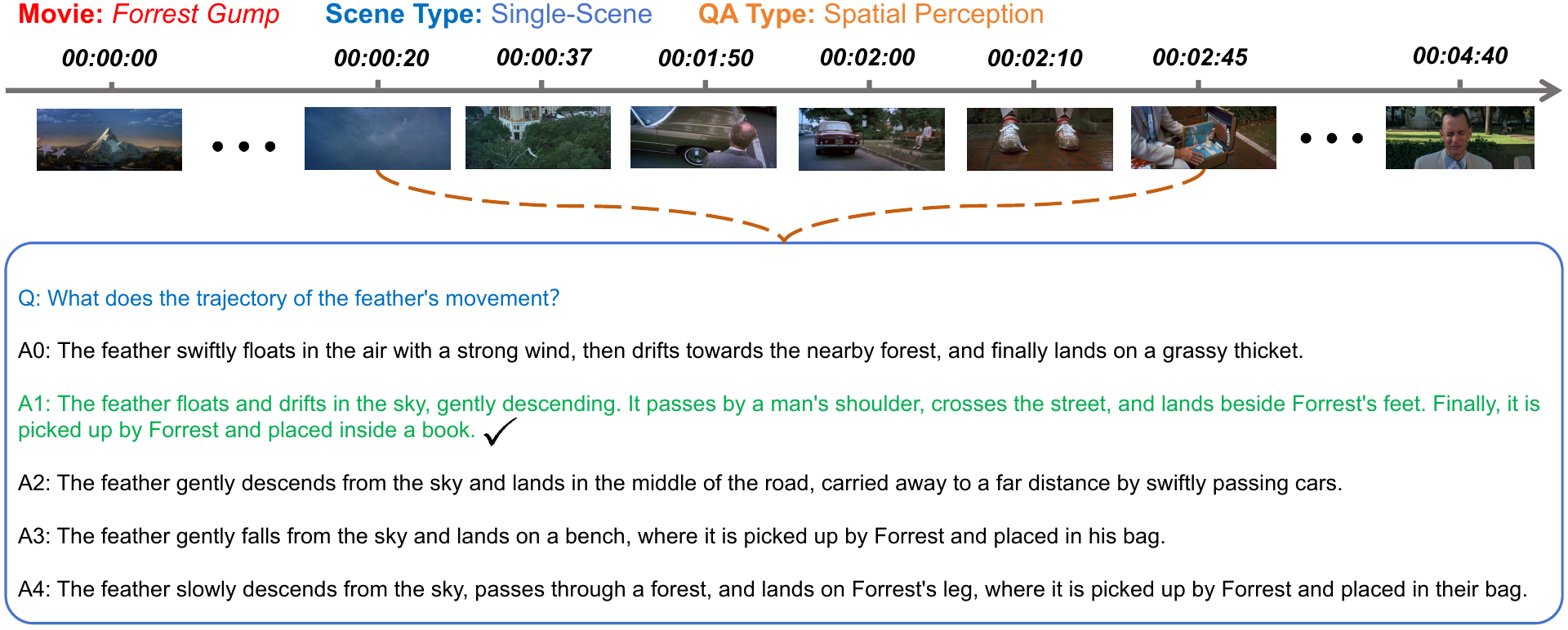}
\end{center}
\vspace{-5mm}
\caption{A Spatial Perception QA example from the \dataset dataset, as well as the prediction from our \method. The ground truth answer is highlighted in green, and the model prediction is indicated by \ding{51}. Best viewed in color.}
\vspace{-2mm}
\label{fig:qa_example3}
\end{figure*}

\begin{figure*}
\begin{center}
\includegraphics[width=0.9\linewidth]{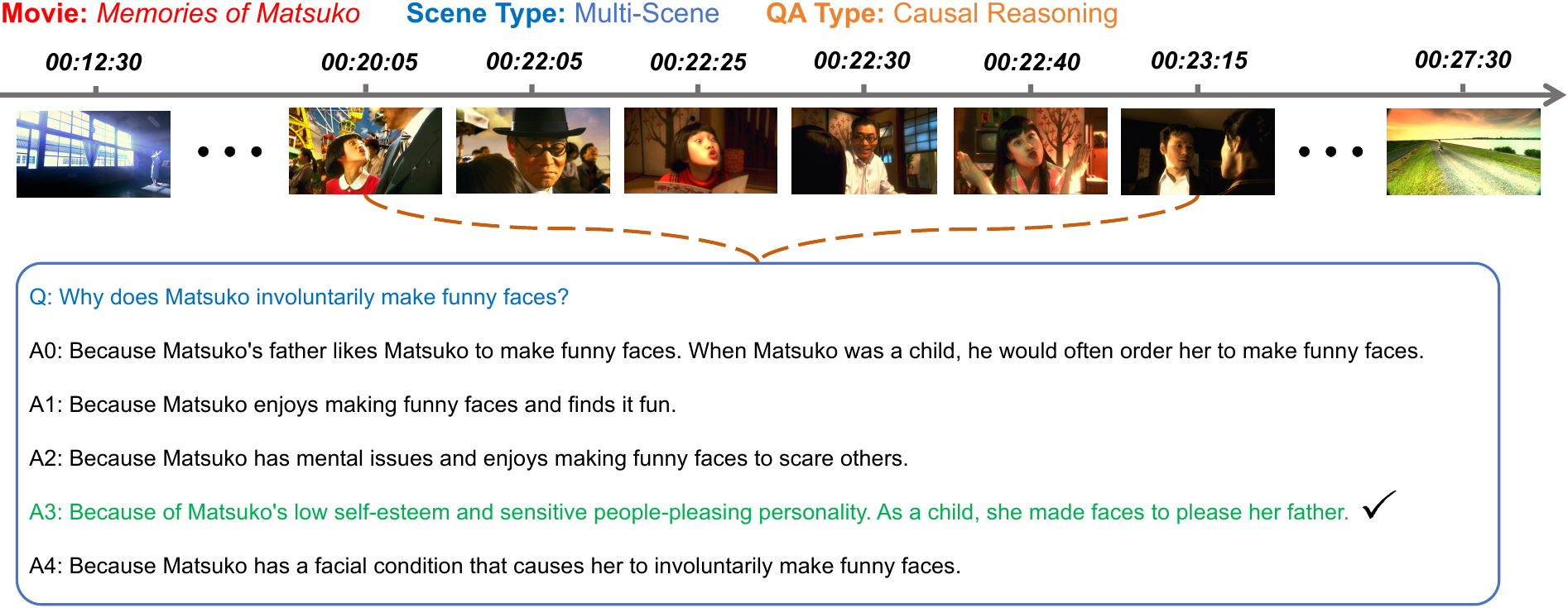}
\end{center}
\vspace{-5mm}
\caption{A Causal Reasoning QA example from the \dataset dataset, as well as the prediction from our \method. The ground truth answer is highlighted in green, and the model prediction is indicated by \ding{51}. Best viewed in color.}
\vspace{-2mm}
\label{fig:qa_example4}
\end{figure*}

\begin{figure*}
\begin{center}
\includegraphics[width=0.9\linewidth]{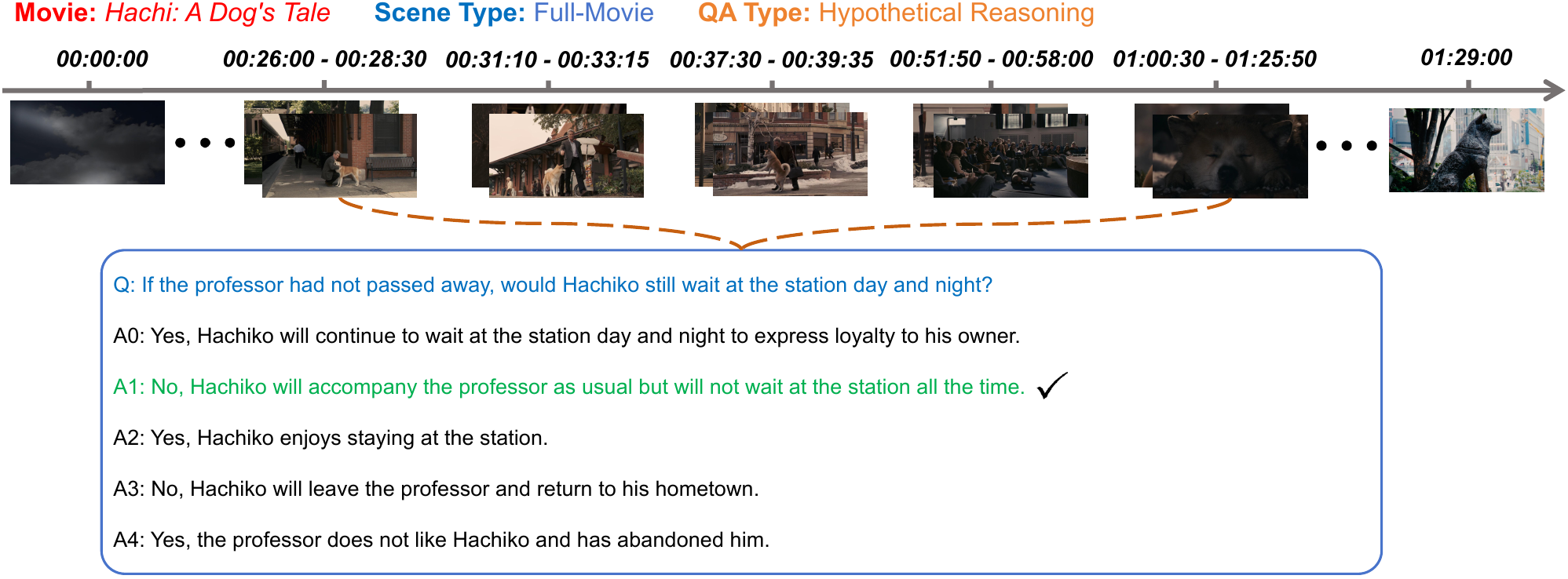}
\end{center}
\vspace{-5mm}
\caption{A Hypothetical Reasoning QA example from the \dataset dataset, as well as the prediction from our \method. The ground truth answer is highlighted in green, and the model prediction is indicated by \ding{51}. Best viewed in color.}
\vspace{-2mm}
\label{fig:qa_example5}
\end{figure*}

\begin{figure*}
\begin{center}
\includegraphics[width=0.9\linewidth]{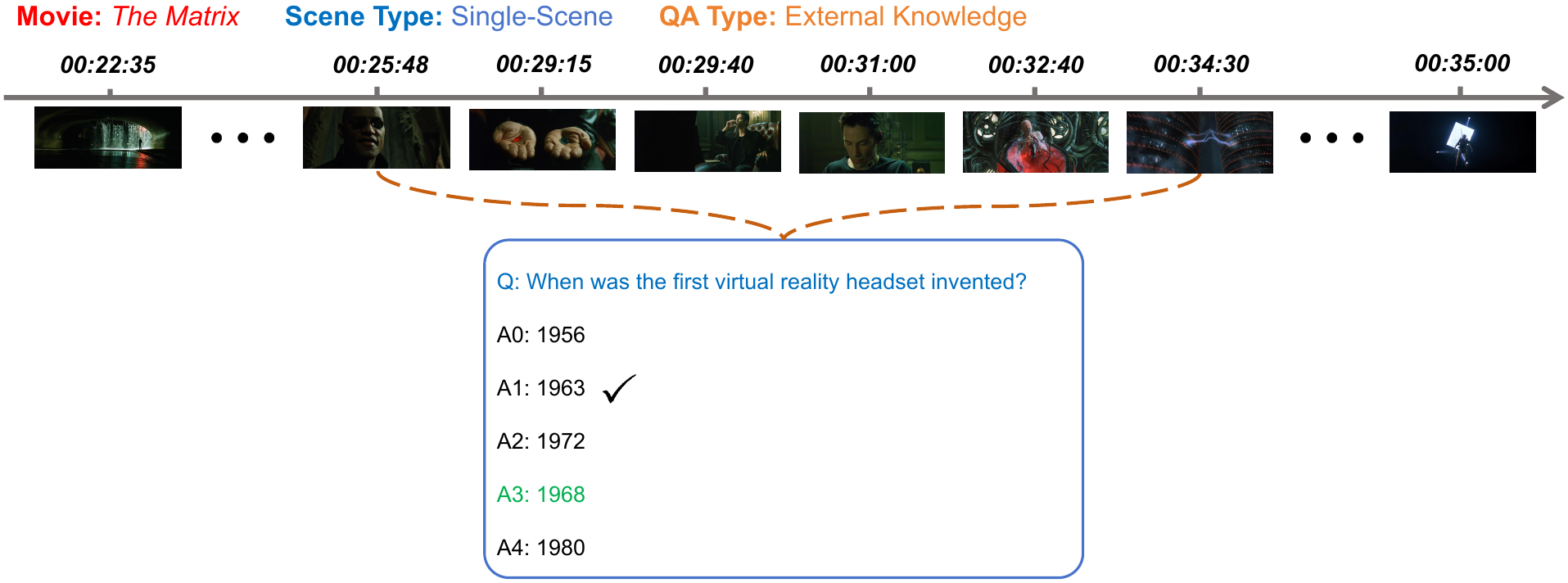}
\end{center}
\vspace{-5mm}
\caption{An External Knowledge QA example from the \dataset dataset, as well as the prediction from our \method. The ground truth answer is highlighted in green, and the model prediction is indicated by \ding{51}. Best viewed in color.}
\vspace{-2mm}
\label{fig:qa_example6}
\end{figure*}

Fig.~\ref{fig:qa_example1}-\ref{fig:qa_example6} show the 6 types of QA tasks designed in the \dataset, which are Information Synopsis, Temporal Perception, Spatial Perception, Causal Reasoning, Hypothetical Reasoning, and External Knowledge. We also present the ground truth answers and the results predicted by our model.

\onecolumn
\begin{center}
\small
\setlength{\tabcolsep}{6pt}
\renewcommand{\arraystretch}{1.2}
\begin{longtable}{|c|p{6cm}|c|p{7.5cm}|}
\caption{100 movies used in our \dataset.} \label{tab:movie_table}\\
\hline
\textbf{ID} & \textbf{Movie Name} & \textbf{Year}  & \textbf{Theme} \\
\hline
\endfirsthead

\hline
\textbf{ID} & \textbf{Movie Name} & \textbf{Year} &  \textbf{Theme} \\
\hline
\endhead

\hline
\multicolumn{4}{r}{\textit{Continued on next page}} \\
\endfoot

\hline
\endlastfoot

1 & The Shawshank Redemption & 1994 & Hope, redemption, friendship \\
2 & Forrest Gump & 1994 & Life journey, destiny, snapshot of American history \\
3 & Farewell My Concubine & 1993 & Tragic love, identity, Peking opera and Cultural Revolution \\
4 & Life Is Beautiful & 1997 & Love and sacrifice, hope through humor in adversity \\
5 & The Legend of 1900 & 1998 & Life choices, solitary genius, art and destiny \\
6 & Schindler's List & 1993 & The Holocaust, sacrifice and redemption \\
7 & Spirited Away & 2001 & Coming-of-age, identity transformation \\
8 & Wall-E & 2008 & Environmentalism, consumerism, love and humanity \\
9 & Titanic & 1997 & Love and tragedy, historical disaster \\
10 & Inception & 2010 & Dream versus reality, the art of inception \\
11 & Three Idiots & 2009 & Friendship, critique of the education system, self-discovery \\
12 & The Chorus & 2004 & Musical inspiration, redemption and hope \\
13 & Hachi: A Dog's Tale & 2010 & Loyalty, unconditional love \\
14 & My Neighbor Totoro & 1988 & Childhood, nature, whimsy \\
15 & The Godfather & 1972 & Family, power, mafia ethics \\
16 & A Chinese Odyssey Part 2: Cinderella & 1994 & Adventure, reincarnation, romance \\
17 & Gone with the Wind & 1939 & Love and survival in wartime, intertwined fate \\
18 & Cinema Paradiso & 1988 & Passion for film, nostalgia, coming-of-age \\
19 & Fight Club & 1999 & Critique of consumerism, identity and rebellion \\
20 & The Truman Show & 1998 & Reality versus simulation, media manipulation and freedom \\
21 & The Lord of the Rings: The Return of the King & 2003 & Epic journey, friendship and sacrifice \\
22 & Roman Holiday & 1953 & Romantic encounter, cross-class love \\
23 & The Cove & 2009 & Environmental protection, animal rights, social activism \\
24 & Lock, Stock and Two Smoking Barrels & 1998 & British gangster tale, luck and ingenuity \\
25 & Intouchables & 2011 & Friendship, overcoming social barriers, human compassion \\
26 & 12 Angry Men & 1957 & Justice, fairness, collective decision-making \\
27 & A Chinese Odyssey Part 1: Pandora's Box & 1994 & Mythology, transformation and romance \\
28 & Flipped & 2010 & First love, growth, change in perspective \\
29 & The Lives of Others & 2006 & Surveillance, oppression, conscience and redemption \\
30 & Amélie & 2001 & Whimsical life, inner transformation, solitary warmth \\
31 & V for Vendetta & 2005 & Totalitarianism and resistance, freedom and revolution \\
32 & Infernal Affairs & 2002 & Dual identities, loyalty and betrayal \\
33 & Scent of a Woman & 1992 & Redemption, mentorship, self-awakening \\
34 & The Dark Knight & 2008 & Order versus chaos, morality and sacrifice \\
35 & The Lord of the Rings: The Two Towers & 2002 & War, loyalty, ensemble heroism \\
36 & The Lord of the Rings: The Fellowship of the Ring & 2001 & Quest, fellowship and courage \\
37 & Silenced & 2011 & Social justice, silenced suffering and awakening \\
38 & A Beautiful Mind & 2001 & Genius, mental struggle, personal triumph and inner life \\
39 & The Godfather Part II & 1974 & Power succession, family and betrayal \\
40 & Edward Scissorhands & 1990 & Alienation, acceptance, societal outcast \\
41 & Se7en & 1995 & Human vices, moral dilemmas, consequences \\
42 & Life of Pi & 2012 & Survival, faith and self-redemption \\
43 & Braveheart & 1995 & Freedom, resistance, national identity \\
44 & The Pianist & 2002 & Survival, trauma, resilience \\
45 & The Prestige & 2006 & Rivalry, obsession, illusion \\
46 & Memories of Matsuko & 2006 & Tragic life, love and redemption \\
47 & Pulp Fiction & 1994 & Non-linear narrative, fate, redemption \\
48 & The Butterfly Effect & 2004 & Time paradox, choices and consequences \\
49 & The Curious Case of Benjamin Button & 2008 & Passage of time, reverse aging and love \\
50 & The Matrix & 1999 & Virtual reality, rebellion, human liberation \\
51 & The Sixth Sense & 1999 & Psychological suspense, life, death and redemption \\
52 & City of God & 2002 & Youth, violence, and corruption in the favelas \\
53 & Eat Drink Man Woman & 1994 & Family relationships, tradition vs. modernity, culinary culture \\
54 & Good Will Hunting & 1997 & Self-discovery, genius, mentorship and personal struggle \\
55 & Grave of the Fireflies & 1988 & Tragedy of war, childhood loss and survival \\
56 & The Monkey King & 2012 & Myth, heroism, and the journey of a legendary figure \\
57 & Almost A Love Story & 1996 & Love, missed opportunities and modern life intersections \\
58 & Chungking Express & 1994 & Urban alienation and ephemeral encounters in a bustling city \\
59 & Legends of the Fall & 1995 & Family bonds, love and transformation in a sweeping epic \\
60 & Pirates of the Caribbean & 2003 & High-seas adventure, piracy and supernatural elements \\
61 & Identity & 2003 & Psychological tension, multiple identities and survival \\
62 & Kikujiro & 1999 & Road trip, unlikely bonding and coming-of-age \\
63 & Shutter Island & 2010 & Sanity, trauma and mind games \\
64 & The Terminal & 2004 & Belonging, cultural clash and bureaucratic absurdity \\
65 & Slumdog Millionaire & 2008 & Destiny, love and hope amid poverty \\
66 & Catch Me If You Can & 2002 & Con artistry, cat-and-mouse pursuit and self-discovery \\
67 & Hotel Rwanda & 2004 & Genocide, survival and moral courage during crisis \\
68 & The Godfather: Part III & 1990 & Family legacy, power struggle and corruption \\
69 & The Attorney & 2013 & Social justice, human rights and political oppression \\
70 & Pride and Prejudice & 2005 & Love, class differences and societal expectations \\
71 & Once A Thief & 1991 & Heists, camaraderie and rogue adventure \\
72 & A Better Tomorrow & 1986 & Brotherhood, honor and loyalty in the criminal underworld \\
73 & A Perfect World & 1993 & Fugitive journey, ethics and unexpected bonds \\
74 & Harry Potter and the Sorcerer's Stone & 2001 & Magic, friendship and the struggle between good and evil \\
75 & Memories of Murder & 2003 & Investigation, human nature and flaws in the justice system \\
76 & The Last Emperor & 1987 & Power, identity and cultural transformation \\
77 & Kekexili: Mountain Patrol & 2004 & Wildlife conservation and social injustice \\
78 & Lord of War & 2005 & Arms dealing, morality and global conflict \\
79 & Hope & 2013 & Tragedy, healing and resilience in the face of loss \\
80 & The Rock & 1996 & Prison escape, military action and heroism \\
81 & Taare Zameen Par & 2007 & Childhood, dyslexia, and the power of nurturing creativity \\
82 & E.T. The Extra-Terrestrial & 1982 & Friendship, wonder and alien encounter \\
83 & The Bourne Identity & 2002 & Identity crisis, espionage and survival \\
84 & Face Off & 1997 & Duality, identity and revenge \\
85 & The Bourne Supremacy & 2004 & Conspiracy, espionage and further identity crisis \\
86 & Days Of Being Wild & 1990 & Existential search, urban alienation and love \\
87 & The Grand Budapest Hotel & 2014 & Eccentricity, nostalgia and meticulously crafted artifice \\
88 & The Man From Earth & 2007 & Philosophical inquiry, immortality and human history \\
89 & In the Mood for Love & 2000 & Forbidden love, longing and subtle intimacy \\
90 & King of Comedy & 1999 & Ambition, media satire and personal dreams \\
91 & The Matrix Revolutions & 2003 & Virtual reality, destiny and the human-machine conflict \\
92 & Triangle & 2009 & Psychological horror, time loops and survival \\
93 & The King's Speech & 2010 & Overcoming adversity, leadership and communication \\
94 & Blood Diamond & 2006 & Conflict, exploitation and ethical dilemmas in war \\
95 & The Terror Live & 2013 & Media ethics, crisis management and public fear \\
96 & Black Hawk Down & 2001 & Combat, military operations and the chaos of war \\
97 & The Mission & 1999 & Underworld dealings, loyalty and betrayal \\
98 & Echoes Of The Rainbow & 2009 & Family, hope and the struggles of everyday life \\
99 & Fist of Legend & 1994 & Honor, revenge and the spirit of martial arts \\
100 & The Warlords & 2007 & Brotherhood, loyalty and political ambition in wartime \\
\end{longtable}
\end{center}
\twocolumn

\clearpage

{\small
\bibliographystyle{spbasic}
\bibliography{egbib}
}
\end{sloppypar}
\end{document}